\documentclass{article}
\usepackage{iclr2027_conference,times}

\usepackage{hyperref}
\usepackage{url}
\usepackage{amsmath,amssymb}
\usepackage{booktabs}
\usepackage{graphicx}
\usepackage{xcolor}
\usepackage{microtype}
\usepackage{multirow}
\usepackage{enumitem}
\usepackage{placeins}
\usepackage{array}
\graphicspath{{figures/}{./}}

\newcommand{\passone}{\textrm{pass@}1}
\newcommand{\execk}[1]{\textrm{exec@}#1}
\newcommand{\oraclek}[1]{\textrm{oracle@}#1}
\newcommand{\GB}{\,\mathrm{GB}}

\newcommand{\codetk}[1]{\textrm{CodeT@}#1}
\newcommand{\sigceil}[1]{\textrm{sig-oracle@}#1}

\title{Fewer Bits, More Tries: Recovering Capability Under Memory Constraints}
\author{Tomer Galanti\\Texas A\&M, Apple \And Josh Susskind\\Apple \And Eran Malach\\Apple}

\iclrfinalcopy % Uncomment for camera-ready version, but NOT for submission.
\begin{document}

\maketitle

\begin{abstract}
Large language models are generally more capable than smaller ones, but their memory requirements can make local deployment difficult. We ask whether part of this capability gap can instead be recovered by running a smaller or more compressed model multiple times and selecting among its outputs using public or model-generated tests. Single-user local inference can leave hardware underutilized, allowing parallel candidate generation to trade additional test-time computation for accuracy at modest latency cost. Across four code-generation benchmarks, verified retry recovers substantial accuracy lost to model size or compression. A 7B Qwen model with eight parallel attempts
essentially matches a 32B model on MBPP+ and exceeds it on the other three benchmarks, while using about $4\times$ less model memory, running up to $6\times$ faster, and using less energy, as measured on Nvidia GPU. We observe similar tradeoff across model sizes, compression levels and hardware choices, although exact benefit depends on the hardware profile and problem setting. These results show that model capacity can be traded for verified test-time computation when correct solutions remain reachable within a modest retry budget.
\end{abstract}

\section{Introduction}
\label{sec:intro}

Deploying a language model under a hard memory budget is usually treated as a compression problem: choose a smaller model, or quantize, prune, or distill a larger one until its weights fit \citep{hinton2015distilling,frantar2023gptq,lin2024awq,frantar2023sparsegpt}. This approach places the entire burden on the compression process: the resulting model must both fit in memory and preserve the accuracy of a larger model. Since language-model performance scales with model size and training compute \citep{kaplan2020scaling,hoffmann2022training}, fitting a smaller checkpoint within a local memory budget can require sacrificing one-shot capability. 

Test-time compute provides another option \citep{snell2025testtime}. Instead of using all available memory for the largest possible model, we can deploy a smaller model and spend more compute on generating and verifying multiple attempts. Repeated sampling and selection are already widely used in code generation and reasoning \citep{chen2021codex,li2022alphacode,wang2023selfconsistency,cobbe2021verifiers,lightman2024verify}, and execution feedback provides a natural way to verify generated programs \citep{ni2023lever,chen2023codet}. Importantly, on local hardware serving a single-user, generating multiple attempts in parallel often results in improved hardware utilization compared to single-stream (batch-size 1) decoding (see detailed discussion in Section~\ref{sec:runtime-analysis}). Consequently, runtime does not increase linearly with the number of parallel attempts, and a moderate number of retries can result in improved accuracy with minimal overhead. We therefore ask the following question:

\textbf{\emph{When can verified attempts from a smaller model substitute for
one attempt from a larger model?}}

The key distinction is between \emph{one-shot accuracy} and \emph{candidate
coverage}: whether a correct solution appears within several attempts. A
smaller or lower-precision model may be less reliable on each draw while still
producing correct solutions after repeated sampling. In that case, verification
can recover much of the apparent capability loss. Across our coding benchmarks,
one-shot accuracy often changes substantially with model size while candidate
coverage changes much less.

Generating a correct candidate is only half the problem; the system must also
recognize it. When external tests are available, they provide a fixed
verification signal. Otherwise, the model must generate its own tests, which
introduces additional ways for retry to fail. We distinguish
\emph{generation failure}, when no correct candidate is produced;
\emph{test-discrimination failure}, when the tests cannot distinguish correct
from incorrect candidates; and \emph{aggregation failure}, when the selection rule chooses the wrong candidate
(for example, failing to discount unreliable tests during majority voting). We analyze these different failure modes across different coding datasets and test-aggregation methods.

{\bf Contributions.\enspace} Our main findings are:
\begin{itemize}[leftmargin=*,itemsep=3pt,topsep=3pt]

\item \textbf{Retry can recover capability lost to smaller models or stronger
compression.}
Across model sizes, quantization levels, and compact model families, much of
the loss in one-shot accuracy can be recovered by verified retry. In a
controlled one-bit experiment, only $2$--$4$ attempts are needed to recover
the bf16 27B model's one-shot accuracy across all four benchmarks while
retaining more than an order-of-magnitude reduction in model memory.

\item \textbf{A smaller model can pay for retry when large-model performance is recovered in few attempts.}
On Nvidia GPUs, Coder-7B with eight attempts essentially matches
Coder-32B with one attempt on MBPP+ and exceeds it on the other three
benchmarks while using about $4.3\times$ less model memory, running
$1.3$--$5.9\times$ faster, and using $1.3$--$6.7\times$ less energy. On Apple Silicon, a smaller model using 2-4 attempts can be faster and more energy-efficient than
Coder-14B.

\item \textbf{Recovery depends on both generating good solutions and recognizing
them.}
Model-generated tests can recover substantial accuracy when external tests are
unavailable, but introduce an additional verification bottleneck. For example, BigCodeBench
is mainly limited by tests that fail to distinguish correct from incorrect
programs, while LiveCodeBench also lacks correct candidates under deep
sampling. While this is admittedly a limitation of our current approach, we believe that using multi-turn interaction with the environment can overcome this verification bottleneck by allowing the model to iteratively debug its solutions and refine its generated tests based on execution feedback.

\end{itemize}

\section{Related work}
\label{sec:related}
% -----------------------------------------------------------------------------

{\bf Test-time scaling and verification.\enspace}
Repeated sampling, aggregation, and verifier-guided selection improve code generation and reasoning
\citep{wang2023selfconsistency,cobbe2021verifiers,uesato2022process,
lightman2024verify,snell2025testtime,brown2024monkeys,zhao2025sample,setlur2025suboptimal};
related work studies candidate coverage, best-of-$N$ selection, and imperfect verification
\citep{gao2023scaling,huang2025bestofn,stroebl2026limits}.
Recent work also shows that an LLM can directly verify and rank its own rollouts
using fine-grained probabilistic scoring \citep{kwok2026llmasaverifiergeneralpurposeverificationframework}.
Execution-grounded consensus further shows that the quality of behavioral
evidence can matter more than the particular aggregation rule
\citep{jiang2026semanticvotingexecutiongroundedconsensus}.
Work closest in deployment motivation jointly allocates scale, precision,
KV-cache memory, and test-time compute or exploits hardware headroom
\citep{kim2026notallbits,hao2026mobiletts}.
We instead characterize when scale or precision can be exchanged for verified
retry and localize the remaining error to generation, test discrimination, or
aggregation.

{\bf Execution-guided code generation.\enspace}
Program synthesis has long used repeated sampling, execution feedback, learned verifiers, and model-generated tests
\citep{austin2021program,chen2021codex,li2022alphacode,le2022coderl,ni2023lever,chen2023codet}.
S* generates distinguishing inputs and DiffCodeGen combines differential testing with behavioral clustering when public tests are unavailable
\citep{li2025sstar,he2026diffcodegen}.
Our contribution is orthogonal to selector design: we hold selectors fixed and study their value under memory constraints.
Verifier-backed committee search similarly separates proposal coverage from candidate recovery
\citep{sunkaraneni2026agentic}.

{\bf Compression and memory-constrained inference.\enspace}
Distillation, pruning, and quantization reduce persistent weight memory
\citep{hinton2015distilling,frantar2023sparsegpt,frantar2023gptq,xiao2023smoothquant,lin2024awq},
while other work studies scale--precision tradeoffs and near-capacity inference
\citep{dettmers2023case4bit,alizadeh2023flash}.
BitCal-TTS examines quantization with adaptive test-time reasoning
\citep{patarlapalli2026bitcal}.
We instead ask how much accuracy verified retry can recover after choosing a checkpoint to satisfy a hard memory budget.
Distilled and pruned models serve as practical deployment alternatives, not controlled descendants of Qwen.

{\bf Serving and hardware utilization.\enspace}
Speculative decoding and serving systems improve inference through drafting, scheduling, batching, and KV-cache management
\citep{leviathan2023speculative,chen2023specsampling,yu2022orca,kwon2023vllm}.
Unlike speculative decoding, which retains the larger target model, retry keeps only the chosen checkpoint resident and spends hardware headroom on additional candidates.
We therefore measure wall clock and energy rather than treating candidate count as normalized compute.

%\section{Theoretical Runtime Analysis}
% \begin{itemize}
%     \item In this section we will show why on local hardware we don't pay a lot for more retries
%     \item Introduce ``roofline model'' or back-of-the-envelope calculation for runtime (IO, compute, and what is the runtime as a function of IO/compute). Give reference in the literature that this roofline model is used all the time.
%     \item Let's assume a network with only MLPs (single $d\times d$ matrix, no up/down projections), hidden dimension $d$. Then, given some hardware profile and batch size, what is the runtime.
%     \item The above explains why single-batch decode is memory bound, and what happens when we do more retries (increase the batch size). How many retries we can add without paying in runtime (or paying very little).
%     \item Why the hardware ``scale'' doesn't matter for the relative runtime, but only the hardware ``profile'' (is that the right word?) - i.e., the ratio between compute and IO. This motivates why we use B200s for most experiments.
%     \item Optional: give a simple theoretical model (maybe just under oracle) for how accuracy grows with number of retries. Then, given some runtime budget, we can estimate the maximal number of retries and derive the predicted accuracy.
%     \item More complex analysis for Transformer - increased compute with number of tokens. Either give a formal analysis or ``hand-wave'' why it doesn't matter that much in some cases.
% \end{itemize}

\section{Why parallel retry can be inexpensive}
\label{sec:runtime-analysis}

Before asking whether retry can recover lost capability, we first ask whether
local hardware can execute additional attempts cheaply. Generating $K$
candidates requires more computation than generating one, but need not require
$K$ times the latency. Batched inference reuses the same model weights across
sequences, allowing additional attempts to exploit otherwise idle compute. We
illustrate this with a simple roofline model
\citep{williams2009roofline,pope2023efficiently}.

{\bf A simple model.\enspace}
Consider a network with $L$ layers, each a $d\times d$ matrix, and let
$P=Ld^2$ be its parameter count. Processing a batch of $B$ sequences requires
approximately $2BP$ FLOPs\footnote{The factor of 2 comes from counting a
multiply and an add as two FLOPs.} while reading $b_wP$ bytes of weights, where
$b_w$ is the number of bytes per weight. Activation traffic scales as
$O(BLd)$ and is smaller than weight traffic when $B\ll d$; we omit it here to
isolate the effect of reusing model weights across the batch. If hardware $H$
has compute throughput $F_H$ and memory bandwidth $\beta_H$, let
$\widehat{\tau}_H(B)$ denote the estimated wall-clock time for one inference
step on a batch of $B$ sequences. Under the roofline approximation, latency is the larger of the time required to move the weights and the time required to perform the computation: $\widehat{\tau}_H(B)
= \max\left\{ \frac{b_wP}{\beta_H}, \frac{2BP}{F_H} \right\}$. Factoring out $b_wP/\beta_H$ gives
\begin{equation}
\widehat{\tau}_H(B)
=
\tfrac{b_wP}{\beta_H}
\max\!\left\{
1,\tfrac{B}{B_\star(H)}
\right\},
\qquad
B_\star(H)
=
\tfrac{b_wF_H}{2\beta_H}.
\label{eq:retry-roofline}
\end{equation}
Here $B_\star(H)$ is the batch size at which the bandwidth and compute terms
are equal. For $B\le B_\star(H)$, inference is bandwidth-bound: the dominant cost is moving the weights, so increasing the batch adds relatively
little latency.  Beyond this point, computation becomes limiting and latency grows approximately
linearly with batch size. For a single prompt with $K$ parallel attempts,
$B\approx K$. More generally, if $N$ prompts are served concurrently,
$B\approx NK$, so existing serving concurrency reduces the available retry
headroom. Thus retry remains inexpensive while $NK\le B_\star(H)$. The
per-prompt memory accounting used later is separate from this serving-level
concurrency.

Equation~\ref{eq:retry-roofline} shows that the relevant hardware quantity is
the compute-to-bandwidth ratio $F_H/\beta_H$. For bf16 weights, $B_\star=F_H/\beta_H$; published NVIDIA specifications give
$B_\star\approx153$ for A100, $B_\star\approx295$ for H100, and $B_\star\approx290$ for
B200, all well above one
\citep{nvidia2021a100,nvidia2022h100,nvidia2026hgxb200}. Thus the retry headroom depends on the hardware
profile rather than its absolute scale: scaling compute and bandwidth together
changes runtime but leaves $B_\star$ unchanged. For Apple Silicon we measure the break-even point empirically.

{\bf Trading model size for retry.\enspace}
Now compare a smaller model with $P_s$ parameters to a reference model with
$P_r>P_s$ on the same hardware and at the same precision. If the reference is
bandwidth-bound at its serving batch size $N$, i.e., $N\le B_\star(H)$, then
\begin{equation}
\tfrac{\widehat{\tau}_{H,s}(NK)}
     {\widehat{\tau}_{H,r}(N)}
\approx
\tfrac{P_s}{P_r}
\max\left\{
1,\tfrac{NK}{B_\star(H)}
\right\}.
\label{eq:retry-scale-cost}
\end{equation}
The smaller model can therefore spend part of its lower inference cost on
additional attempts. In this toy model, it remains no slower than the reference
while
$K \le K_{\rm sys} := \frac{P_r}{P_s}\frac{B_\star(H)}{N}$.
Reducing numerical precision similarly decreases weight traffic, although
low-precision kernels can also change effective compute throughput.

{\bf From systems headroom to capability recovery.\enspace}
Let $K_{\rm rec}$ be the smallest retry budget at which the smaller or
compressed model recovers the reference model's one-shot accuracy. The useful
regime is therefore $K_{\rm rec}\le K_{\rm sys}$. If recovery occurs before the systems advantage is exhausted, a smaller model
can match the larger model while remaining faster and using less memory.
Otherwise, retry may preserve the memory advantage but lose the latency
advantage.

The roofline model is deliberately approximate: modern Transformers add
attention, KV-cache traffic, sequence-length effects, and serving overhead, and
it implicitly assumes attempts have similar lengths. Length variation can
increase latency when a batch waits for its slowest sequence; adaptive stopping
could mitigate this effect but is outside the fixed-retry setting studied here.
We use the model to identify the expected dependence on model size, retry depth,
and hardware profile rather than to predict measured speedups quantitatively;
the actual break-even points are measured on NVIDIA GPUs and Apple Silicon.
%Modern Transformers add attention, KV-cache traffic, sequence-length effects,
%and implementation-specific overheads, so the calculation is not intended to
%predict measured speedups quantitatively. Its role is to identify the basic
%mechanism and the expected dependence on model size, retry depth, and hardware
%profile; we measure the actual break-even points on NVIDIA GPUs and Apple
%Silicon.

\section{Main results with public-test verification}
\label{sec:main-results}

Section~\ref{sec:runtime-analysis} suggests that retry is useful when
$K_{\rm rec}\lesssim K_{\rm sys}$. We first test this regime using
benchmark-visible tests, covered in this section; Section~\ref{sec:codet-delivery} then considers
model-generated verification.

\subsection{Experimental setting}
\label{sec:setup}

{\bf Deployment configurations.\enspace}
A deployment consists of a checkpoint, retry budget $K$, and serving configuration (e.g., serving engine and batching setup). The model generates $K$ candidates in parallel, and the verifier
selects one using only deployment-time information; correctness is evaluated on
held-out tests. We denote public-test selected accuracy by \execk{$K$} and define
$K_{\rm rec}$ as the smallest tested $K$ that matches or exceeds the reference
model's one-shot accuracy.

{\bf Models and compression.\enspace}
Our main model-size experiments use the Qwen2.5-Coder-Instruct family, spanning
1.5B to 32B parameters \citep{hui2024qwen25coder}. The primary scaling
comparisons use the native bf16 checkpoints. To study precision separately
from model size, we evaluate the 3B, 7B, and 32B models using 16-, 8-, and
4-bit representations and multiple quantization schemes, including GPTQ and
AWQ.

We complement this with a controlled extreme-compression experiment using
bf16, 4-bit, and one-bit versions of the same Qwen3.6-27B model. Because the
architecture and parameter count are fixed, this comparison isolates the
effect of precision. Additional model families and a sparse mixture-of-experts
replication appear in Appendix~\ref{app:deployment-evidence}.

{\bf Benchmarks.\enspace}
We evaluate four code-generation benchmarks. HumanEval+ contains 164 tasks and
MBPP+ 378 tasks, using EvalPlus \citep{liu2023evalplus}. We use 175 problems
from LiveCodeBench \texttt{release\_v5} dated on or after 2024-08-01
\citep{jain2025livecodebench}. For BigCodeBench
\citep{zhuo2025bigcodebench}, we remove tasks whose reference solution fails
under our evaluation pipeline, leaving 1,034 of the original 1,140 tasks for
the main Qwen2.5-Coder experiments. 

{\bf Public-test verification.\enspace}
For EvalPlus, the verifier can use the benchmark's base tests; for
LiveCodeBench, it can use the public examples. BigCodeBench does not provide a
predefined public/private test split, so for each task we deterministically
partition its tests into visible tests used for selection and held-out tests
used only for final evaluation.

For each candidate, we execute the visible tests and select the candidate with
the highest mean token log-probability among those that pass. If none passes,
we return the candidate with the highest mean token log-probability overall.
Thus, the fallback is used only when the visible tests provide no passing
candidate. Final correctness is always determined using held-out tests.

{\bf Candidate generation.\enspace}
Unless stated otherwise, candidates are sampled with temperature $0.8$ and
top-$p=0.95$. The main accuracy experiments generate 50 candidates per task.
Results at smaller retry budgets are computed from these fixed pools, so
changes with $K$ reflect additional attempts rather than different generation
runs. For each task and retry budget $K$, we compute the exact expected
accuracy under a uniformly sampled size-$K$ subset of the 50 candidates,
using the selector's ranking of the candidates; the full pool is used at
$K=50$.

{\bf Memory accounting.\enspace}
Model memory is checkpoint weights plus KV cache, computed from realized prompt
and completion lengths with the shared prompt counted once. We report measured
process memory separately in Appendix Table~\ref{tab:memory-calibration}.

{\bf Runtime and energy.\enspace}
Runtime and energy are measured directly for the checkpoint and retry budget
being served rather than inferred from parameter count or candidate count.
Our main GPU measurements use vLLM \citep{kwon2023vllm} on NVIDIA B200 GPUs
with fixed workloads of 40--60 prompts. Multiple candidates are decoded in
parallel, so the measured wall time captures the batching effect described in
Section~\ref{sec:runtime-analysis}.

The B200 provides a controlled high-throughput platform on which to measure
the retry curve, but it is not intended to stand in for every local device (see the note below).
We therefore also run two physical-device studies on 64-GiB
M3 Max and M5 Pro systems using llama.cpp. The first compares one-bit
Qwen3.6-27B with its bf16 parent, while the second compares smaller dense
Qwen2.5-Coder models with Coder-14B at their benchmark-specific $K_{\rm rec}$,
the smallest retry budget that recovers the reference model's one-shot
accuracy. These measurements probe the same memory--retry tradeoff under a
substantially different hardware and serving regime.

For each comparison, we report wall-clock speedup and energy saving relative
to the reference deployment: $\text{speedup} = \frac{T_{\rm ref}}{T_{\rm dep}}$ and $\text{energy saving} = \frac{E_{\rm ref}}{E_{\rm dep}}$, so values above one favor the smaller or compressed deployment. Memory saving
is defined analogously.

{\bf Statistical comparisons.\enspace}
Accuracy differences are computed per task, with 95\% confidence intervals
estimated from 10,000 bootstrap resamples of tasks \citep{efron1979bootstrap}. When comparing one-shot and retry, both quantities
are computed from the same underlying candidate pool, preserving the task-level
pairing in each resample.

\emph{{\bf A note on our hardware choice.\enspace} While our work is motivated by the deployment of local models, many of our experiments are conducted on Nvidia B200 GPUs, high-end hardware rarely used by individual consumers. However, as discussed in Section \ref{sec:runtime-analysis}, the runtime speedup due to multiple retries depends on the \textbf{hardware profile} (the ratio between compute and memory throughput) rather than absolute throughput numbers. The Nvidia B200, designed for large-scale training, is compute-optimized, scaling compute throughput far more than memory bandwidth. Our B200 results are therefore complemented by experiments on current Apple Silicon, which offers a more balanced hardware profile. Ultimately, our results should be viewed as a tool to drive the discussion on design choices for future hardware, rather than a simple benchmarking of current devices.}

\subsection{Results}
\label{sec:public-results}

\begin{figure*}[t]
\centering
{\setlength{\tabcolsep}{2pt}%
\begin{tabular}{@{}cccc@{}}
\multicolumn{4}{c}{
  \includegraphics[width=\textwidth]{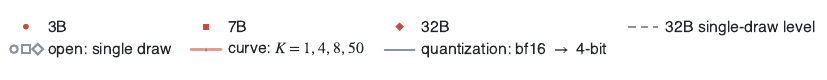}
} \\[-4pt]

\includegraphics[width=0.235\textwidth]{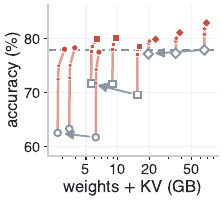} &
\includegraphics[width=0.235\textwidth]{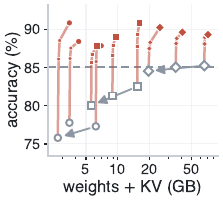} &
\includegraphics[width=0.235\textwidth]{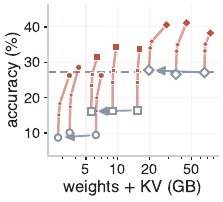} &
\includegraphics[width=0.235\textwidth]{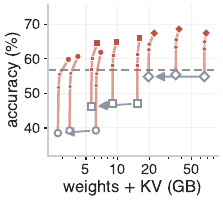}
\\[-2pt]

\small (a) MBPP+ &
\small (b) HumanEval+ &
\small (c) LiveCodeBench &
\small (d) BigCodeBench
\end{tabular}}

\caption{\textbf{Retry trades a small amount of additional memory for higher accuracy.}
Each curve traces accuracy from $K=1$ to $K=50$ against model memory, defined as
checkpoint weights plus KV cache. The dashed line marks the one-shot accuracy of
native bf16 Coder-32B.}
\label{fig:accuracy-memory-concrete}
\end{figure*}

Figure~\ref{fig:accuracy-memory-concrete} gives the overall memory--accuracy
tradeoff. Increasing the retry budget substantially improves accuracy while
adding relatively little KV-cache memory compared with the difference in
checkpoint size. We next ask whether this accuracy can be recovered without
giving up the smaller model's runtime advantage.

\begin{table*}[t]
\centering
\scriptsize
\setlength{\tabcolsep}{2.0pt}
\renewcommand{\arraystretch}{1.05}
\caption{\textbf{Eight attempts from a smaller model can match or exceed one attempt from Coder-32B at lower serving cost.}
The 3B, 7B, and 14B models use $K=8$, while Coder-32B uses $K=1$.
Accuracy comes from the main candidate pools; wall time and energy are measured
on matched B200 workloads. Speedup and energy saving are Coder-32B/retried
model, so values above $1\times$ favor the smaller model; the Coder-32B row is
the reference. See Appendix Table \ref{tab:scale-draw-surface-complete} for more details.}
\label{tab:scale-draw-surface}
\resizebox{\textwidth}{!}{%
\begin{tabular}{@{}l rrr rrr rrr rrr@{}}
\toprule
& \multicolumn{3}{c}{\textbf{MBPP+}}
& \multicolumn{3}{c}{\textbf{HumanEval+}}
& \multicolumn{3}{c}{\textbf{LiveCodeBench}}
& \multicolumn{3}{c}{\textbf{BigCodeBench}} \\
\cmidrule(lr){2-4}\cmidrule(lr){5-7}\cmidrule(lr){8-10}\cmidrule(l){11-13}
Checkpoint
& Acc. & \shortstack{Wall\\speedup} & \shortstack{Energy\\saving}
& Acc. & \shortstack{Wall\\speedup} & \shortstack{Energy\\saving}
& Acc. & \shortstack{Wall\\speedup} & \shortstack{Energy\\saving}
& Acc. & \shortstack{Wall\\speedup} & \shortstack{Energy\\saving} \\
\midrule
Coder-3B ($K{=}8$)
& 74.2\% & $2.8\times$ & $3.5\times$
& 87.5\% & $2.2\times$ & $2.7\times$
& 19.8\% & $4.0\times$ & $5.8\times$
& 58.2\% & $1.5\times$ & $1.7\times$ \\
\bfseries Coder-7B ($K{=}8$)
& \bfseries 77.8\% & \bfseries $2.0\times$ & \bfseries $1.9\times$
& \bfseries 89.4\% & \bfseries $2.2\times$ & \bfseries $2.4\times$
& \bfseries 27.5\% & \bfseries $5.9\times$ & \bfseries $6.7\times$
& \bfseries 62.7\% & \bfseries $1.3\times$ & \bfseries $1.3\times$ \\
Coder-14B ($K{=}8$)
& 79.4\% & $1.1\times$ & $1.0\times$
& 89.5\% & $1.0\times$ & $0.9\times$
& 33.9\% & $1.5\times$ & $1.5\times$
& 66.3\% & $0.6\times$ & $0.6\times$ \\
Coder-32B ($K{=}1$)
& 77.9\% & $1.0\times$ & $1.0\times$
& 85.1\% & $1.0\times$ & $1.0\times$
& 27.1\% & $1.0\times$ & $1.0\times$
& 56.6\% & $1.0\times$ & $1.0\times$ \\
\bottomrule
\end{tabular}}
\end{table*}

\begin{figure*}[t]
\centering
{\setlength{\tabcolsep}{2pt}%
\begin{tabular}{@{}cccc@{}}
\multicolumn{4}{c}{
  \includegraphics[width=\textwidth]{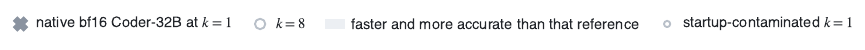}
} \\[-4pt]
\includegraphics[width=0.235\textwidth]{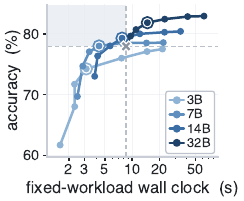} &
\includegraphics[width=0.235\textwidth]{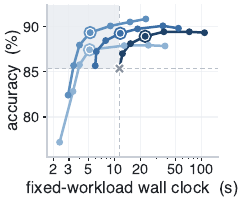} &
\includegraphics[width=0.235\textwidth]{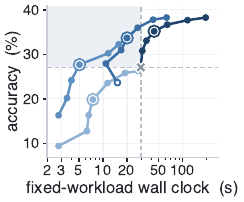} &
\includegraphics[width=0.235\textwidth]{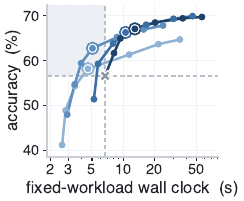}
\\[-2pt]
\small (a) MBPP+ &
\small (b) HumanEval+ &
\small (c) LiveCodeBench &
\small (d) BigCodeBench
\end{tabular}}

\caption{\textbf{Parallel retry can make a smaller model both faster and more accurate.}
The curves show accuracy versus measured B200 wall-clock time as the number of
attempts increases. The $\times$ marks Coder-32B with one attempt, and the
circles mark eight attempts.}
\label{fig:accuracy-time}
\end{figure*}

\begin{table*}[t]
\centering
\small
\setlength{\tabcolsep}{5.0pt}
\renewcommand{\arraystretch}{1.14}

\caption{\textbf{A few retries recover the one-shot accuracy lost to one-bit quantization.}
Each row compares the Bonsai Q1\_0 one-bit checkpoint with the bf16 version of
the same 27B model. ``Tries to recover'' is the smallest tested number of
attempts for which the one-bit model matches or exceeds bf16 one-shot accuracy.
Wall-time and energy ratios are bf16@1 / one-bit@recovery, so values above one
favor the one-bit deployment and values below one favor bf16.}
\label{tab:bonsai-onebit}

\resizebox{\textwidth}{!}{%
\begin{tabular}{@{}lccccccc@{}}
\toprule
& \multicolumn{4}{c}{\textbf{Accuracy}}
& \multicolumn{3}{c}{\textbf{Cost of recovering bf16 one-shot accuracy}} \\
\cmidrule(lr){2-5}
\cmidrule(l){6-8}

Benchmark
& \shortstack{One-shot\\1-bit / bf16}
& \shortstack{At recovery\\1-bit / bf16}
& \shortstack{Selected after 50 tries\\1-bit / bf16}
& \shortstack{Any correct in 50\\1-bit / bf16}
& \shortstack{Tries to\\recover}
& \shortstack{Wall-time\\ratio}
& \shortstack{Energy\\ratio} \\
\midrule

HumanEval+
& 85.5\% / 89.3\%
& 91.5\% / 89.3\%
& 93.9\% / 93.9\%
& 98.2\% / 98.2\%
& $2$
& $0.72\times$
& $1.30\times$ \\

MBPP+
& 68.9\% / 77.2\%
& 79.8\% / 77.2\%
& 82.5\% / 83.6\%
& 87.3\% / 88.6\%
& $4$
& $0.42\times$
& $0.73\times$ \\

LiveCodeBench
& 14.6\% / 16.8\%
& 20.9\% / 16.8\%
& 34.9\% / 34.3\%
& 35.4\% / 34.3\%
& $2$
& $1.04\times$
& $1.61\times$ \\

BigCodeBench
& 19.2\% / 26.7\%
& 29.9\% / 26.7\%
& 63.4\% / 67.0\%
& 72.3\% / 70.8\%
& $2$
& $0.71\times$
& $0.92\times$ \\

\bottomrule
\end{tabular}%
}
\end{table*}

{\bf A smaller model can substitute parallel retry for model scale.\enspace}
We begin with the central deployment question: can a model that is substantially
smaller in memory use its lower inference cost to generate additional candidates
and recover the accuracy of a much larger model without becoming slower? On
B200, the answer is yes over a broad operating range.

Table~\ref{tab:scale-draw-surface} compares native Qwen2.5-Coder models using
eight parallel attempts with Coder-32B using a single attempt. Coder-7B at
$K=8$ essentially matches Coder-32B at $K=1$ on MBPP+ and exceeds it on the
other three benchmarks. At the same time, it runs $1.3$--$5.9\times$ faster
and uses $1.3$--$6.7\times$ less energy. Its model memory---checkpoint weights
plus KV cache---is about $4.3\times$ smaller, and its measured B200 peak
process memory is $21.58\GB$ rather than $73.02\GB$, a $3.38\times$ reduction
(Appendix Table~\ref{tab:memory-calibration}).

This is the favorable regime identified in Section~\ref{sec:runtime-analysis}:
the accuracy lost by reducing model scale is recovered before parallel retry
exhausts the smaller model's systems advantage. Figure~\ref{fig:accuracy-time}
shows the same phenomenon across retry budgets. Initially, increasing $K$
moves the smaller models primarily upward in accuracy while adding relatively
little wall time; at larger $K$, latency grows more rapidly and the gains begin
to saturate.

The effect is not simply that any small model becomes competitive after enough
sampling. Coder-3B at $K=8$, for example, remains below the 32B reference on
MBPP+ and LiveCodeBench, while Coder-14B at $K=8$ is more accurate but already
loses the latency and energy advantage on BigCodeBench. The useful operating
point therefore depends on how quickly accuracy recovers relative to the
systems cost of increasing $K$. We note that while we force a fixed $K$ throughout our experiments, choosing $K$ adaptively based on the problem setting is another path for improving performance, which we leave to future work.

{\bf Recovery is possible when compression reduces reliability more than
reachability.\enspace}
Why can a much smaller or more compressed model recover so quickly? The key
quantity is not only one-shot accuracy, but whether correct solutions remain
reachable after repeated sampling.

The controlled Qwen3.6-27B experiment isolates this effect: architecture
and parameter count are fixed, and only numerical precision changes. One-bit
quantization lowers one-shot accuracy by $2.2$--$8.3$ percentage points across
the four benchmarks, while the fraction of tasks solved at least once within
50 attempts changes by at most $1.5$ points. As a result, only $K=2$--$4$
verified attempts are needed to recover the bf16 model's one-shot accuracy on
every benchmark (Table~\ref{tab:bonsai-onebit}), so compression increases
$K_{\rm rec}$ only modestly.

Reducing model scale can behave differently. On BigCodeBench, Coder-7B and
Coder-32B differ by $8.7$ percentage points on one-shot accuracy, but by only
$1.4$ points in the fraction of tasks solved within 50 attempts. Much of the
32B advantage is therefore a reliability advantage that retry can recover.
LiveCodeBench provides the counterexample: Coder-7B solves $5.7$ percentage
points fewer tasks within 50 attempts than Coder-32B. Across the full native
scale ladder, candidate coverage varies by $13.1$ points on LiveCodeBench,
compared with only $1.3$--$2.4$ points on the other three benchmarks
(Appendix~\ref{app:oracle-coverage}). Figure~\ref{fig:memory-routes} in the appendix summarizes this contrast. Here reducing model capacity removes reachable solutions, so additional sampling cannot fully recover the loss.

This distinction is also visible under conventional 4- and 8-bit
quantization. Across the direct bf16-to-lowest-memory comparisons, weight memory falls by roughly $2.7$--$3.6\times$, while accuracy after 50 verified attempts changes by at most $3.1$ percentage points (Appendix Table~\ref{tab:precision-direct}). Together, these results suggest that retry is most effective when reducing memory primarily lowers the probability of success on each draw rather than eliminating successful outputs from the
model's support.

{\bf The runtime advantage is largest at moderate retry depth.\enspace}
Preserving candidate coverage is necessary but not sufficient:
$K_{\rm rec}$ must also remain within the retry budget that the hardware can
execute efficiently. Figure~\ref{fig:accuracy-time} shows this tradeoff.
On MBPP+, Coder-7B obtains $83\%$ of its eventual $K=50$ accuracy gain by
$K=4$ and $93\%$ by $K=8$, while wall time increases by only $1.35\times$ and $1.73\times$, respectively (Appendix Table~\ref{tab:retry-operating-points}). Beyond this range, accuracy
gains diminish while wall time continues to rise.

This is consistent with the roofline picture: retry is inexpensive while batching headroom remains, but eventually compute, KV-cache traffic, or serving overhead dominates. The key question is whether recovery occurs before this systems break-even point.

{\bf Recovery depth predicts whether retry pays off, but the break-even $K$ is
device-specific.\enspace}
Our two physical-device studies on Apple Silicon (64\,GiB M3 Max and M5 Pro,
served with \texttt{llama.cpp}) test this prediction on a substantially
different hardware and serving stack than the B200 results above.
Appendix Table~\ref{tab:apple-main} shows the same dependence on recovery depth for the
dense model-size comparisons. On both M3 Max and M5 Pro, every measured
deployment that recovers the Coder-14B one-shot accuracy within at most four
attempts is faster and more energy-efficient than the 14B reference. Coder-7B
requires only two attempts on HumanEval+, MBPP+, and BigCodeBench and four on
LiveCodeBench, and all four comparisons remain favorable on both chips.

The boundary becomes visible for Coder-3B. HumanEval+ and BigCodeBench recover
at $K=4$ and remain favorable on both devices. MBPP+ requires $K=8$:
the deployment is clearly slower and more energy-intensive on M3 Max, while
the M5 Pro measurements lie near wall-time break-even and remain unfavorable
in energy. LiveCodeBench requires $K=16$ and is unfavorable in both wall time
and energy on both chips. The memory advantage remains substantial throughout,
at roughly $4.3\times$ for the 3B deployments, but deeper recovery eventually
spends the latency and energy advantage.

The controlled one-bit experiment shows that the break-even point is not a
universal value of $K$. At the same fixed $K=4$, Bonsai Q1\_0 is slower than its
bf16 parent on M3 Max across all four workloads, with speedup
$0.83$--$0.87\times$, but is $1.58$--$1.67\times$ faster on M5 Pro.
The same deployment therefore falls on opposite sides of latency break-even
on the two devices.

This cross-device reversal is qualitatively consistent with the hardware
dependence in Section~\ref{sec:runtime-analysis}: $K_{\rm sys}$ is determined
by the interaction between the model and the hardware's compute--bandwidth
profile, not by retry depth alone. The roofline calculation is not intended to
predict the measured ratios quantitatively---kernel efficiency, memory
capacity, serving software, and other device-specific effects also matter---
but it correctly anticipates that the same $K_{\rm rec}$ can be inexpensive
on one system and too costly on another.

\section{Model-generated verification}
\label{sec:codet-delivery}

The previous section assumes that reliable tests are available at deployment
time. We now consider the harder setting in which they are not. In this case,
the model must generate not only candidate solutions, but also the evidence
used to choose among them. This makes verification itself part of the
capability that may be lost when the model is made smaller or more compressed.

\subsection{Experimental setting}

{\bf Model-generated tests.\enspace}
Given the task specification, the model generates a set of candidate tests.
We execute every candidate solution on these tests and select an answer using
CodeT \citep{chen2023codet}. Rather than simply choosing the candidate that
passes the most tests, CodeT exploits patterns of agreement between candidates
and generated tests to identify groups of behaviorally consistent solutions.
If no usable generated tests are produced, we fall back to the candidate with
the highest mean token log-probability. Final correctness is always evaluated
with separate held-out tests, exactly as in the public-test setting.

{\bf Diagnosing verification failure.\enspace}
A failed selection can have three distinct causes. First, the candidate pool
may contain no correct solution. We measure this with \oraclek{$K$}, the
fraction of tasks for which at least one of the $K$ candidates is correct.

Second, the generated tests may produce identical outcomes for a correct
candidate and an incorrect one. We call the pattern of tests that a candidate
passes or fails its \emph{test-outcome signature}. Candidates with the same
signature cannot be distinguished using these test results alone. We define
\sigceil{$K$} as the accuracy of an oracle selector that chooses optimally
among the signatures observed in the $K$ candidates, but cannot distinguish
candidates that share a signature. This gives an upper bound on selection
accuracy when the selector relies only on the generated-test outcomes.

Third, the selector may fail to exploit the information contained in those test
outcomes. For example, a selection rule that does not properly discount unreliable tests will fail to choose a correct solution, even though the tests do, technically, distinguish correct from incorrect solutions. The gap from CodeT to \sigceil{$K$} measures this aggregation failure;
the gap from \sigceil{$K$} to \oraclek{$K$} measures test-discrimination
failure; and the remaining error above \oraclek{$K$} measures candidate
generation failure.

\subsection{Results}

\begin{table*}[t]
\centering
\scriptsize
\setlength{\tabcolsep}{2.4pt}
\renewcommand{\arraystretch}{1.08}

\caption{\textbf{Model-generated tests let a compact 7B model recover much of
the 32B model's one-shot accuracy.}
We report Coder-7B AWQ4 with model-generated verification at $K=8$ and $K=50$,
alongside public-test selection and the Coder-32B bf16 one-shot reference.
``vs.\ 32B'' is the accuracy difference from that reference.}
\label{tab:codet-compact}

\resizebox{\textwidth}{!}{%
\begin{tabular}{@{}lrrrrrrr rrr@{}}
\toprule
& \multicolumn{7}{c}{\textbf{Accuracy}}
& \multicolumn{3}{c}{\textbf{Memory}} \\
\cmidrule(lr){2-8}
\cmidrule(l){9-11}

Benchmark
& \shortstack{7B\\one-shot}
& \shortstack{7B, 8 tries\\generated tests}
& \shortstack{vs.\ 32B\\one-shot}
& \shortstack{7B, 50 tries\\generated tests}
& \shortstack{vs.\ 32B\\one-shot}
& \shortstack{7B, 50 tries\\public tests}
& \shortstack{32B\\one-shot}
& \shortstack{Weights (GB)\\7B / 32B}
& \shortstack{Memory\\saving}
& \shortstack{Process peak (GB)\\7B / 32B} \\
\midrule

MBPP+
& 71.6\% & 76.8\% & $-1.1$ & 78.6\% & $+0.7$
& 79.9\% & 77.9\%
& 5.57 / 65.53
& $10.3\times$
& 11.73 / 73.02 \\

HumanEval+
& 80.0\% & 82.9\% & $-2.2$ & 83.5\% & $-1.6$
& 87.8\% & 85.1\%
& 5.57 / 65.53
& $10.4\times$
& 11.73 / 73.02 \\

LiveCodeBench
& 16.1\% & 25.0\% & $-2.1$ & 28.0\% & $+0.9$
& 31.4\% & 27.1\%
& 5.57 / 65.53
& $10.3\times$
& 11.73 / 73.02 \\

BigCodeBench
& 46.3\% & 52.7\% & $-3.9$ & 53.5\% & $-3.2$
& 67.9\% & 56.6\%
& 5.57 / 65.53
& $10.3\times$
& 11.73 / 73.02 \\

\bottomrule
\end{tabular}%
}
\end{table*}

{\bf Model-generated verification recovers substantial capability.\enspace}
Table~\ref{tab:codet-compact} shows that the memory--retry tradeoff survives
even when external tests are removed. Coder-7B AWQ4 improves substantially
over its one-shot accuracy on all four benchmarks. With $K=50$, model-generated
verification reaches $78.6\%$, $83.5\%$, $28.0\%$, and $53.5\%$ on MBPP+,
HumanEval+, LiveCodeBench, and BigCodeBench, respectively. These values are
within $3.2$ percentage points of the displayed Coder-32B one-shot reference
on every benchmark, while the 7B checkpoint uses about $10\times$ less model
memory and $6.23\times$ less measured process memory.

Much of the benefit appears at moderate retry depth: with only eight attempts,
the 7B model is already within $1.1$--$3.9$ percentage points of the 32B
one-shot reference across the four benchmarks. The effect is also not limited
to this particular checkpoint. Across 58 verified model--benchmark
configurations on MBPP+, HumanEval+, and BigCodeBench, CodeT improves one-shot
accuracy by a median of $7.3$ percentage points; the middle half of the gains
range from $3.3$ to $11.4$ points.

\begin{figure*}[t]
\centering
{\setlength{\tabcolsep}{2pt}%
\begin{tabular}{@{}cccc@{}}
\multicolumn{4}{c}{
  \includegraphics[width=\textwidth]{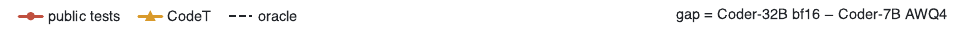}
} \\[-4pt]
\includegraphics[width=0.235\textwidth]{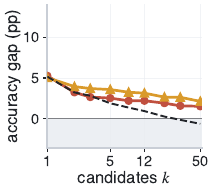} &
\includegraphics[width=0.235\textwidth]{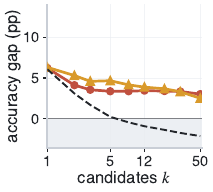} &
\includegraphics[width=0.235\textwidth]{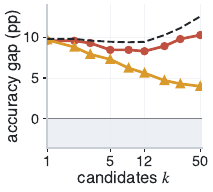} &
\includegraphics[width=0.235\textwidth]{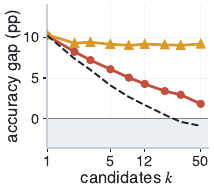}
\\[-2pt]
\small (a) HumanEval+ &
\small (b) MBPP+ &
\small (c) LiveCodeBench &
\small (d) BigCodeBench
\end{tabular}}

\caption{\textbf{Retry closes much of the gap between the 7B and 32B models when the smaller model still generates correct solutions.}
The curves show the accuracy difference between Coder-32B bf16 and Coder-7B
AWQ4 as the number of attempts increases, using public tests, model-generated
tests (CodeT), and the fraction of tasks with at least one correct candidate.
The gap largely disappears on HumanEval+, MBPP+, and BigCodeBench, but remains
on LiveCodeBench because the 7B model fails to generate correct solutions for
some tasks.}
\label{fig:gap-closes}
\end{figure*}

{\bf Generated verification remains substantially weaker than external
tests.\enspace}
Table~\ref{tab:codet-compact} also shows that generating candidates and
recognizing them are different problems. On BigCodeBench, for example, public-test selection reaches
$67.9\%$ at $K=50$, whereas model-generated verification reaches only
$53.5\%$. To understand where this gap comes from, we increase the candidate
budget to $K=500$ and hold the candidate pools and generated-test outcomes
fixed while comparing the selector with the two diagnostic upper bounds.

Appendix Figure~\ref{fig:ceiling} reveals different bottlenecks across benchmarks.
At $K=500$, the gap from CodeT to the signature oracle is between $4.3$ and $7.7$ percentage points across the coding benchmarks. Thus, some information already present in the
generated-test outcomes is not fully exploited by the selector. More importantly, even the signature oracle remains below candidate coverage
by $5.4-18.6$ percentage points. BigCodeBench is the
clearest example: many correct and incorrect programs produce identical
outcomes on the generated tests, so no selector using only those tests can
reliably distinguish them. The main limitation there is therefore not merely
the aggregation rule, but the information provided by the generated tests
themselves.

LiveCodeBench exhibits a different failure mode. Even after 500 attempts,
$57\%$ of tasks contain no correct candidate at all. For these tasks, neither
better generated tests nor a better selector can help. Model-generated
verification is therefore constrained by both sides of the retry problem:
the generator must continue to produce correct solutions, and the generated
tests must contain enough information to recognize them.

These results sharpen the conclusion of the public-test experiments.
Parallel retry can recover substantial capability even without an external
verifier, but model-generated verification introduces an additional bottleneck. When
correct candidates remain reachable, improving the quality of generated tests
can be as important as increasing the number of attempts.

% \section{Main Results (with public-tests)}
% Motivation, connect it to the high lever motivation and theoretical analysis (i.e., we convinced you that in theory running more times makes sense, now we will show you this in practice)
% \subsection{Experimental Settings}
% (not necessarily in this order)
% \begin{itemize}
%     \item How do we measure memory, runtime and energy
%     \item What are the ``deployment'' configuration parameters
%     \item How do we analyze accuracy under public-tests
%     \item Benchmarks details
%     \item model details
%     \item Generation details
%     \item Apple silicon (here or in a separate section)
% \end{itemize}

% \subsection{Results (Nvidia GPU)}
% Main results of the paper (in order of how important they are?)
% \subsection{Results on Apple silicon}
% Explain why we suspet that the results are different than what we could predict from Nvidia GPU (different hardware profile, llama.cpp, wrong back-of-the-envelope because hardware can heat)

% \section{Model generated tests}
% \subsection{Experimental setting}
% \subsection{Results}

% \subsection{Discussion, Conclusion, limitation and Future work}

\section{Discussion and Conclusion}
\label{sec:discussion}

Interest in running language models locally keeps growing, but a gap remains
between local and cloud deployment, due to strict compute and memory limitations: the
largest checkpoint that fits on a laptop is weaker than the one in a datacenter.
This treats local inference as a restricted version of cloud inference, and so local models will always remain inferior to those served on the cloud.

We believe this misses the main advantage of local hardware: we do not share it with anyone. A datacenter GPU is kept busy
by many users, so every extra token generated by the model competes with demand from other users. A personal
device serving one user is mostly idle, and single-stream decoding leaves most
of its compute unused. Therefore, given a single user request, generating multiple attempts can be done almost ``for
free.'' In this paper we showed that this free compute buys back much of the
capability lost to a smaller model: a 7B model with eight parallel
attempts matches or exceeds a 32B model with one, while using less memory, less
time, and less energy. This works because a smaller model is still likely to generate a correct solution given enough attempts.

We note that our results are limited to single-turn code generation and depend on having good
tests, which the model cannot always write for itself. A natural next step is extending this method to an interactive, multi-turn, local coding agent, which already writes and runs tests and already spends
most of its time waiting on one sequential stream. This opens up questions on
how such a harness should spend the idle capacity: on parallel solutions, on
parallel plans, or on better verification.
Another promising direction is moving beyond single-candidate selection by actively combining parallel generations or allowing concurrent processes to communicate and refine their outputs (with a similar approach as the one studied in \cite{park2026scaling}). Finally, models could be explicitly trained or distilled not for single-shot performance, but to maximize aggregate success across multiple parallel attempts under local hardware constraints.

% Our results suggest that the key question is not only how much one-shot
% accuracy is lost when a model is made smaller or more compressed, but whether
% correct solutions remain reachable. Quantization often preserves this reachability even when one-shot accuracy drops, allowing a small number of
% verified retries to recover the loss. Reducing model scale can instead remove
% solutions from the model's support, as on LiveCodeBench, where additional
% sampling cannot fully close the gap.

% Whether retry is useful in practice also depends on verification and systems
% cost. Public tests provide a strong selection signal, while model-generated
% tests can fail to distinguish correct from incorrect candidates. At the same
% time, retry is attractive only while the required recovery depth remains within
% the hardware's efficient batching regime. The best memory-constrained
% deployment is therefore not necessarily the largest model that fits, but the
% one whose recovery depth remains affordable on the target hardware.

% Our experiments are limited to code generation and a small set of hardware and
% serving stacks, and our self-verification systems measurements do not include
% the cost of generating and executing tests. Still, the results establish a
% simple regime in which model capacity can be exchanged for verified test-time
% computation: when correct solutions remain reachable and only a modest number
% of attempts is needed to recover them.

\newpage

% \section*{AI Use Statement}

% We used generative AI tools, to assist with manuscript drafting and editing, literature review, conceptual framing, organization and presentation, software development and debugging, and discussion of experimental design and the interpretation of results. AI-assisted suggestions were treated as input to the research process rather than as authoritative outputs: the authors reviewed the resulting code, analyses, text, and scientific claims against the underlying experiments and data. Generative AI was not used as a source of experimental data or evidence. The authors take full responsibility for the final content of the paper, including all AI-assisted text, code, analyses, and claims.

% \section*{Reproducibility Statement}

% We provide the experimental setup, model and benchmark configurations, candidate
% generation and verification procedures, and statistical methodology in
% Section~\ref{sec:setup}. The appendix gives additional details on task
% populations, memory accounting, process-memory calibration, runtime and energy
% measurement, candidate--verifier matching, and execution validation
% (Appendix~\ref{app:repro}). It also reports the full precision sweeps and
% additional deployment comparisons underlying the main results. Code and
% experiment artifacts will be provided as supplementary material.

\bibliography{references}
\bibliographystyle{iclr2027_conference}

\newpage
\appendix

\section{Additional experimental results}
We add here a more complete version of the tables appearing in the main text.
\begin{table*}[t]
\centering
\scriptsize
\setlength{\tabcolsep}{2.0pt}
\renewcommand{\arraystretch}{1.05}
\caption{\textbf{Eight attempts from a smaller model can match or exceed one attempt from Coder-32B at lower serving cost.}
The 3B, 7B, and 14B models use $K=8$, while Coder-32B uses $K=1$.
Accuracy comes from the main candidate pools; wall time and energy are measured
on matched B200 workloads. Speedup and energy saving are Coder-32B/retried
model, so values above $1\times$ favor the smaller model.}
\label{tab:scale-draw-surface-complete}
\resizebox{\textwidth}{!}{%
\begin{tabular}{@{}l rrrrr rrrrr rrrrr rrrrr@{}}
\toprule
& \multicolumn{5}{c}{\textbf{MBPP+} \; ($77.9\%$ ref.)}
& \multicolumn{5}{c}{\textbf{HumanEval+} \; ($85.1\%$ ref.)}
& \multicolumn{5}{c}{\textbf{LiveCodeBench} \; ($27.1\%$ ref.)}
& \multicolumn{5}{c}{\textbf{BigCodeBench} \; ($56.6\%$ ref.)} \\
\cmidrule(lr){2-6}\cmidrule(lr){7-11}\cmidrule(lr){12-16}\cmidrule(l){17-21}
Checkpoint
& Acc. & Wall & \shortstack{Wall\\speedup} & J/task & \shortstack{Energy\\saving}
& Acc. & Wall & \shortstack{Wall\\speedup} & J/task & \shortstack{Energy\\saving}
& Acc. & Wall & \shortstack{Wall\\speedup} & J/task & \shortstack{Energy\\saving}
& Acc. & Wall & \shortstack{Wall\\speedup} & J/task & \shortstack{Energy\\saving} \\
\midrule
Coder-3B ($K{=}8$)
& 74.2\% & 3.12 & $2.8\times$ & 32  & $3.5\times$
& 87.5\% & 5.17 & $2.2\times$ & 60  & $2.7\times$
& 19.8\% & 7.46 & $4.0\times$ & 120 & $5.8\times$
& 58.2\% & 4.59 & $1.5\times$ & 54 & $1.7\times$ \\
\bfseries Coder-7B ($K{=}8$)
& \bfseries 77.8\% & \bfseries 4.31 & \bfseries $2.0\times$ & \bfseries 59  & \bfseries $1.9\times$
& \bfseries 89.4\% & \bfseries 5.25 & \bfseries $2.2\times$ & \bfseries 68  & \bfseries $2.4\times$
& \bfseries 27.5\% & \bfseries 5.06 & \bfseries $5.9\times$ & \bfseries 104 & \bfseries $6.7\times$
& \bfseries 62.7\% & \bfseries 5.07 & \bfseries $1.3\times$ & \bfseries 72 & \bfseries $1.3\times$ \\
Coder-14B ($K{=}8$)
& 79.4\% & 7.75 & $1.1\times$ & 110 & $1.0\times$
& 89.5\% & 11.66 & $1.0\times$ & 175 & $0.9\times$
& 33.9\% & 20.08 & $1.5\times$ & 450 & $1.5\times$
& 66.3\% & 10.51 & $0.6\times$ & 157 & $0.6\times$ \\
Coder-32B ($K{=}1$)
& 77.9\% & 8.57 & $1.0\times$ & 112 & $1.0\times$
& 85.1\% & 11.41 & $1.0\times$ & 162 & $1.0\times$
& 27.1\% & 29.82 & $1.0\times$ & 692 & $1.0\times$
& 56.6\% & 6.69 & $1.0\times$ & 92 & $1.0\times$ \\
\bottomrule
\end{tabular}}
\end{table*}

% =============================================================================
\section{Additional evidence for the memory--retry tradeoff}
\label{app:deployment-evidence}
% =============================================================================

The Qwen3.6-27B experiments study extreme one-bit compression, while the Qwen2.5-Coder experiments cover more conventional quantization and changes in model size. We complement the accuracy results with wall-clock and energy measurements on B200 GPUs and Apple silicon, together with process-memory measurements on B200. Throughout, model memory refers to checkpoint weights plus the KV cache required for generation; measured process-level peaks are reported separately in Table~\ref{tab:memory-calibration}.

% -----------------------------------------------------------------------------
\subsection{Extreme one-bit compression and preserved candidate coverage}
\label{app:bonsai}

We compare bf16, 4-bit, and one-bit versions of the same Qwen3.6-27B model to isolate the effect of numerical precision. We also evaluate model-generated verification to test whether the one-bit model can recover lost one-shot accuracy when it must generate the tests used to select among its own candidates.

Table~\ref{tab:bonsai-full} reports the accuracy and model-generated verification results. All accuracy measures are evaluated on the full task population for each experiment.

\begin{table*}[!t]
\centering
\scriptsize
\setlength{\tabcolsep}{4.0pt}
\renewcommand{\arraystretch}{1.06}

\caption{\textbf{Accuracy and model-generated verification for the Qwen3.6-27B precision variants.}
All accuracy columns use the full task population for each experiment.
``Usable tests'' reports the fraction of tasks for which a usable generated
test suite was produced.}
\label{tab:bonsai-full}

\begin{tabular}{@{}llrrrrrr@{}}
\toprule
Benchmark
& Checkpoint
& \passone
& \execk{8}
& \execk{50}
& \oraclek{50}
& \codetk{8}
& \shortstack{Usable\\tests} \\
\midrule

\multirow{3}{*}{HumanEval+}
& Bonsai Q1\_0 (1-bit) & 85.5\% & 93.3\% & 93.9\% & 98.2\% & 91.2\% & 0.707 \\
& Bonsai AWQ4 (4-bit)  & 89.9\% & 94.1\% & 94.5\% & 98.2\% & 93.1\% & 0.866 \\
& Qwen3.6-27B (bf16)   & 89.3\% & 93.6\% & 93.9\% & 98.2\% & 93.3\% & 0.927 \\

\midrule
\multirow{3}{*}{MBPP+}
& Bonsai Q1\_0 (1-bit) & 68.9\% & 81.0\% & 82.5\% & 87.3\% & 78.2\% & 0.810 \\
& Bonsai AWQ4 (4-bit)  & 76.6\% & 82.3\% & 83.3\% & 87.8\% & 80.8\% & 0.918 \\
& Qwen3.6-27B (bf16)   & 77.2\% & 83.5\% & 83.6\% & 88.6\% & 82.1\% & 0.899 \\

\midrule
\multirow{3}{*}{LiveCodeBench}
& Bonsai Q1\_0 (1-bit) & 14.6\% & 28.7\% & 34.9\% & 35.4\% & 37.2\% & 0.977 \\
& Bonsai AWQ4 (4-bit)  & 15.5\% & 29.4\% & 34.3\% & 34.3\% & 37.1\% & 0.514 \\
& Qwen3.6-27B (bf16)   & 16.8\% & 30.5\% & 34.3\% & 34.3\% & 38.1\% & 0.754 \\

\midrule
\multirow{3}{*}{BigCodeBench}
& Bonsai Q1\_0 (1-bit) & 19.2\% & 51.5\% & 63.4\% & 72.3\% & 40.4\% & 0.835 \\
& Bonsai AWQ4 (4-bit)  & 23.9\% & 54.0\% & 67.3\% & 72.9\% & 30.5\% & 0.935 \\
& Qwen3.6-27B (bf16)   & 26.7\% & 53.9\% & 67.0\% & 70.8\% & 33.8\% & 0.962 \\

\bottomrule
\end{tabular}
\end{table*}

{\bf Reliability versus reachability.\enspace}
One-bit compression hurts the probability of solving a task on any single attempt much more than it changes whether the model can solve the task at all after repeated sampling. On HumanEval+, Q1\_0 loses $3.8$ percentage points in \passone{}, yet both Q1\_0 and bf16 solve $98.2\%$ of tasks at least once within 50 attempts. On MBPP+, the one-shot gap is $8.3$ percentage points, while the corresponding gap after 50 attempts is only $1.3$ points. The same pattern appears on BigCodeBench: \passone{} falls from $26.7\%$ to $19.2\%$, but the fraction of tasks solved at least once within 50 attempts remains within $1.5$ percentage points of bf16. LiveCodeBench is noisier, with Q1\_0 slightly exceeding bf16 at deeper retry; we treat these small differences as sampling variation rather than evidence that one-bit compression improves the model.

The 4-bit checkpoint is generally close to bf16, although the ordering is not perfectly monotone with precision. The robust pattern is instead that one-bit compression substantially reduces single-attempt reliability while leaving the set of tasks that can be solved after repeated attempts much less affected.

{\bf LiveCodeBench formatting failures.\enspace}
LiveCodeBench also exposes a separate failure mode. Only about one quarter of generations contain the fenced code block expected by the evaluation pipeline: $25.0\%$, $25.8\%$, and $26.5\%$ for Q1\_0, AWQ4, and bf16, respectively. Among generations with the expected format, held-out pass rates are $58.3\%$, $60.2\%$, and $63.2\%$; generations without a fenced block never pass. Moreover, 78, 91, and 93 of the 175 tasks produce no correctly formatted candidate at all within 50 attempts. The low LiveCodeBench scores therefore reflect not only problem-solving ability, but also a formatting failure shared across the model family rather than one caused specifically by one-bit compression. We evaluate the models under the fixed deployed pipeline rather than conditioning on generations that happened to satisfy the expected format.

{\bf Model-generated verification.\enspace}
BigCodeBench shows that generating good candidates and selecting among them are
not the same problem. On the full Bonsai evaluation population, \codetk{8}
reaches $40.4\%$, $30.5\%$, and $33.8\%$ for Q1\_0, AWQ4, and bf16,
respectively. Thus, one-bit compression substantially lowers one-shot accuracy
while still benefiting strongly from model-generated verification.

LiveCodeBench reveals a different limitation. A compact line-based format
produces usable generated tests for $97.7\%$, $51.4\%$, and $75.4\%$ of tasks
for Q1\_0, AWQ4, and bf16, compared with only $0.6\%$, $1.1\%$, and $2.3\%$
under the earlier JSON-array format at the same 5,120-token budget. The
improvement comes from preserving completed tests before generation is
truncated: $133/175$, $146/175$, and $131/175$ tasks still reach the token
limit.

Despite substantial differences in test-generation success across checkpoints,
\codetk{8} is similar for Q1\_0, AWQ4, and bf16 ($37.20\%$, $37.12\%$, and
$38.13\%$). Because failed test generation falls back to confidence ranking,
these scores include the cost of test-generation failure rather than
conditioning on successful test suites.

{\bf How many attempts are needed to recover the parent?\enspace}
Table~\ref{tab:bonsai-onebit} defines recovery as the smallest tested value of
$K$ for which Q1\_0 matches or exceeds the bf16 model's one-shot accuracy.
HumanEval+, LiveCodeBench, and BigCodeBench recover at $K=2$. MBPP+ requires
$K=4$, although the boundary is extremely close: at $K=2$, Q1\_0 reaches
$77.1981\%$, compared with $77.2381\%$ for bf16, a difference of only
$0.04$ percentage points. At $K=4$, Q1\_0 reaches $79.83\%$. We therefore
report $K=4$ as the observed recovery point, while noting that the distinction
between two and four attempts on MBPP+ is very small.

On BigCodeBench, the recovery is much clearer. Q1\_0 improves from $19.22\%$
with one attempt to $29.93\%$ with two, exceeding the bf16 one-shot accuracy of
$26.70\%$. These intermediate-$K$ accuracies are computed from the existing
$K=50$ candidate pools.

{\bf Systems cost of recovery.\enspace}
Recovering accuracy is useful only if the additional attempts do not erase the
systems advantage of the compressed model. We first compare Q1\_0 and bf16 at
the same value of $K$ on HumanEval+ to measure how efficiently each checkpoint
uses the serving stack.

\begin{table*}[!t]
\centering
\scriptsize
\setlength{\tabcolsep}{3.8pt}
\renewcommand{\arraystretch}{1.06}

\caption{\textbf{Q1\_0 is faster and more energy-efficient than bf16 at the same retry budget.}
Results use the HumanEval+ serving workload. Entries compare Q1\_0 with bf16 at
the same $K$. Speedup and energy saving are bf16/Q1\_0, while throughput ratio
is Q1\_0/bf16. Values are medians over three repetitions.}
\label{tab:bonsai-systems}

\resizebox{\textwidth}{!}{%
\begin{tabular}{@{}rcccccccc@{}}
\toprule
$K$
& \shortstack{Weights (GB)\\Q1 / bf16}
& \shortstack{Weight\\saving}
& \shortstack{Wall (s)\\Q1 / bf16}
& \shortstack{Wall-clock\\speedup}
& \shortstack{tok/s\\Q1 / bf16}
& \shortstack{Throughput\\ratio}
& \shortstack{J/task\\Q1 / bf16}
& \shortstack{Energy\\saving} \\
\midrule

1
& 3.80 / 54.66
& $14.38\times$
& 271.7 / 359.7
& $1.32\times$
& 222.8 / 156.5
& $1.42\times$
& 1{,}656.2 / 3{,}946.8
& $2.38\times$ \\

8
& 3.80 / 54.66
& $14.38\times$
& 1{,}549.1 / 2{,}719.3
& $1.76\times$
& 307.4 / 166.0
& $1.85\times$
& 10{,}367 / 30{,}382
& $2.93\times$ \\

50
& 3.80 / 54.66
& $14.38\times$
& 8{,}603.8 / 16{,}477.5
& $1.92\times$
& 346.8 / 171.1
& $2.03\times$
& 59{,}151 / 185{,}108
& $3.13\times$ \\

\bottomrule
\end{tabular}}
\end{table*}

Table~\ref{tab:bonsai-systems} shows that at equal $K$, Q1\_0 is
consistently faster and uses less energy than bf16.
Its advantage also grows with the retry budget: wall-clock speedup increases
from $1.32\times$ at $K=1$ to $1.92\times$ at $K=50$, and energy saving
increases from $2.38\times$ to $3.13\times$. This shows that the one-bit
checkpoint can use wide parallel batches efficiently. It does not, however,
tell us whether compression is cheaper after accounting for the additional
attempts needed to recover accuracy.

For that comparison, we run Q1\_0 at the benchmark-specific recovery budget and
compare it with bf16 at $K=1$. The result depends strongly on the workload.
Wall-clock speedup and energy saving are
$0.72\times$ and $1.30\times$ on HumanEval+,
$0.42\times$ and $0.73\times$ on MBPP+,
$1.04\times$ and $1.61\times$ on LiveCodeBench, and
$0.71\times$ and $0.92\times$ on BigCodeBench.
Thus, the memory advantage remains large, but recovery does not always save
time or energy. On MBPP+, for example, the additional attempts make Q1\_0
slower and more energy-intensive than bf16, whereas on LiveCodeBench it is
slightly faster and uses substantially less energy.

We repeated the HumanEval+ and BigCodeBench recovery measurements in both
execution orders to check for order effects. The two wall-time/energy ratios are
$0.675/1.289$ and $0.762/1.312$ on HumanEval+, and
$0.7079/0.9068$ and $0.7191/0.9288$ on BigCodeBench. The main table reports
the geometric mean across the two orders. Energy is more stable than wall time
across these repetitions, so we avoid attaching significance to small
differences in the latency ratios.

{\bf Model-memory accounting for the Apple comparison.\enspace}
For the fixed-$K=4$ Bonsai comparison in Table~\ref{tab:apple-main}, model
memory is the checkpoint weights plus the KV cache implied by the realized
prompt and completion lengths. Mean prompt lengths are $216.6$, $96.3$,
$646.1$, and $363.0$ tokens on HumanEval+, MBPP+, LiveCodeBench, and
BigCodeBench, respectively. Relative to bf16 at $K=1$, the resulting memory
savings are $13.41\times$, $13.45\times$, $13.31\times$, and $11.43\times$.
The smaller saving on BigCodeBench is due to its substantially longer
completions.

Qwen3.6-27B uses a hybrid architecture with 48 linear-attention layers and 16
full-attention layers. Only the full-attention layers require a KV cache that
grows with sequence length, corresponding to $65{,}536$ bytes per token. The
recurrent state of the linear-attention layers does not grow with sequence
length and is therefore excluded from Equation~\ref{eq:kv-memory}. We also
verified the prompt accounting independently: after normalizing llama.cpp's
\texttt{prompt\_tokens} counter by the number of tasks and parallel slots, it
matches the separately tokenized prompt lengths. We therefore count the shared
prompt once and each generated completion separately.

{\bf BigCodeBench extraction.\enspace}
Code extraction matters for the Qwen3.6/Bonsai BigCodeBench results. Without
\texttt{evalplus.sanitize}, the regex fallback can select the wrong code block.
On the bf16 pool, canonical sanitization raises the parse rate from $72.9\%$ to
$93.1\%$ and \passone{} from $9.18\%$ to $26.70\%$ on the same generations.
For the non-reasoning Qwen2.5-Coder models, the canonical and legacy extraction
procedures agree up to whitespace on all 120 checked examples, so their BigCodeBench results are unchanged. All Qwen3.6/Bonsai results use canonical
sanitized extraction.

{\bf Serving concurrency.\enspace}
For the matched serving sweeps, each checkpoint uses the fastest slot count
found in its tuning sweep. Q1\_0 uses 16 slots at $K=8$ and $K=50$, while
bf16 uses 4. The bf16 $K=50$ configuration was not exhaustively reswept over
every possible slot count, so the reported throughput ratios should be read as
comparisons between the best configurations we measured rather than as globally
optimized serving points.
% -----------------------------------------------------------------------------
\subsection{Physical-device measurements on Apple Silicon}
\label{app:apple-fleet}

\begin{table*}[t]
\centering
\scriptsize
\setlength{\tabcolsep}{2.2pt}
\renewcommand{\arraystretch}{0.94}

\caption{\textbf{Physical-device recovery depends on retry depth.}
Accuracy is deployment/reference. Memory saving, speedup, energy saving, and
power ratio are reference/deployment. In the model-scale block, semicolon-separated
entries follow the order of the two comparisons in the second column.
Model memory is checkpoint weights plus KV cache from realized sequence lengths.}
\label{tab:apple-main}

\resizebox{\textwidth}{!}{%
\begin{tabular}{@{}llccrrrrrrrr@{}}
\toprule
& & & &
\multicolumn{4}{c}{\textbf{M3 Max, 64 GiB}} &
\multicolumn{4}{c}{\textbf{M5 Pro, 64 GiB}} \\
\cmidrule(lr){5-8}
\cmidrule(l){9-12}

Benchmark
& Comparison
& \shortstack{Accuracy\\dep./ref.}
& \shortstack{Memory\\saving ($\times$)}
& \shortstack{Speedup\\($\times$)}
& \shortstack{Energy\\saving ($\times$)}
& \shortstack{Power\\ratio ($\times$)}
& Swap
& \shortstack{Speedup\\($\times$)}
& \shortstack{Energy\\saving ($\times$)}
& \shortstack{Power\\ratio ($\times$)}
& Swap \\
\midrule

\multicolumn{12}{@{}l}{\textit{One-bit quantization}} \\

HumanEval+
& Q1@4 / 27B@1
& 93.1/89.3
& 13.41
& 0.85
& 0.88
& 1.04
& yes/no
& 1.67
& 1.06
& 0.63
& yes/no \\

MBPP+
& Q1@4 / 27B@1
& 79.8/77.2
& 13.45
& 0.86
& 0.83
& 0.97
& yes/no
& 1.59
& 1.05
& 0.66
& yes/no \\

LiveCodeBench
& Q1@4 / 27B@1
& 25.6/16.8
& 13.31
& 0.87
& 0.87
& 1.00
& yes/no
& 1.66
& 1.08
& 0.65
& yes/no \\

BigCodeBench
& Q1@4 / 27B@1
& 41.4/26.7
& 11.43
& 0.83
& 0.82
& 0.99
& yes/no
& 1.58
& 1.03
& 0.65
& yes/no \\

\midrule
\multicolumn{12}{@{}l}{\textit{Reducing model scale}} \\

HumanEval+
& 7B@2 / 14B@1; 3B@4 / 14B@1
& 85.7/85.6; 85.9/85.6
& 1.94; 4.34
& 2.14; 2.21
& 2.18; 2.04
& 1.04; 1.00
& no/no; no/no
& 2.36; 2.44
& 2.23; 2.01
& 0.95; 0.83
& no/no; no/no \\

MBPP+
& 7B@2 / 14B@1; 3B@8 / 14B@1
& 74.7/73.1; 74.3/73.1
& 1.94; 4.33
& 2.12; 0.76
& 2.04; 0.75
& 0.96; 0.99
& no/no; no/no
& 2.15; 0.98
& 2.05; 0.80
& 0.96; 0.82
& no/no; no/no \\

BigCodeBench
& 7B@2 / 14B@1; 3B@4 / 14B@1
& 53.5/48.8; 52.2/48.8
& 1.94; 4.32
& 1.78; 2.05
& 1.82; 1.92
& 1.00; 0.93
& no/no; no/no
& 1.77; 2.43
& 1.68; 1.87
& 0.95; 0.77
& no/no; no/no \\

LiveCodeBench
& 7B@4 / 14B@1; 3B@16 / 14B@1
& 24.2/23.6; 23.6/23.6
& 1.94; 4.14
& 1.46; 0.53
& 1.50; 0.63
& 1.05; 1.22
& no/no; no/no
& 1.66; 0.68
& 1.68; 0.79
& 1.02; 1.16
& no/no; no/no \\

\bottomrule
\end{tabular}%
}
\end{table*}

We run two matched experiments with llama.cpp on 64-GiB M3 Max and M5 Pro
systems. The first compares Qwen3.6-27B bf16 at $K=1$ with Bonsai Q1\_0 at
$K=4$. The second compares fp16 GGUF Qwen2.5-Coder-3B and Coder-7B with
Coder-14B, using the retry budget required for the smaller model to recover the
14B model's one-shot accuracy. Within each comparison, the reference and
retried configurations are measured on the same physical device using
deterministic eight-prompt batches. We record generation wall time, completion
tokens, throughput, package power, energy, and swap activity. The only
exception is M3 Max BigCodeBench in the Bonsai comparison, where two disjoint
task shards are measured on configuration-matched M3 Max systems.

We report energy only when every expected batch has a power record, power
sampling covers at least half of each batch, and the measurements pass our
consistency checks. For Bonsai, the two chips give qualitatively different
results. On M3 Max, Q1\_0 at $K=4$ is slower than bf16 on all four workloads,
with wall-clock speedups of $0.83$--$0.87\times$ and energy-saving ratios of
$0.82$--$0.88\times$. On M5 Pro, it is $1.58$--$1.67\times$ faster, with
energy-saving ratios of $1.03$--$1.08\times$.

The difference is also visible in throughput. Q1\_0 reaches
$3.48$--$3.57\times$ the bf16 throughput on M3 Max and
$6.60$--$6.85\times$ on M5 Pro. Since Q1\_0 generates four candidates for
each bf16 candidate, the M5 Pro throughput gain is large enough to offset the
additional work, whereas the M3 Max gain is not. The bf16 model uses swap in
every matched Bonsai comparison, while Q1\_0 does not. We do not attribute the
cross-chip reversal to swap or to any single hardware mechanism; the
measurements establish the serving outcome rather than its unique cause.

The dense-model comparison shows a clear dependence on recovery depth.
Coder-7B recovers Coder-14B one-shot accuracy with two attempts on HumanEval+,
MBPP+, and BigCodeBench, and with four attempts on LiveCodeBench. Every measured
$K\leq4$ comparison is favorable in both wall time and energy on both chips.
Coder-3B at $K=8$ on MBPP+ is clearly unfavorable on M3 Max; on M5 Pro its two
wall-time ratios, $1.03\times$ and $0.92\times$, straddle break-even, while
its energy-saving ratios, $0.85\times$ and $0.76\times$, remain below one.
At $K=16$ on LiveCodeBench, Coder-3B is unfavorable on both chips, with
wall-clock speedup/energy saving of $0.53\times/0.63\times$ on M3 Max and
$0.68\times/0.79\times$ on M5 Pro. Thus recovery depth predicts the broad
systems regime, but the precise margin depends on the hardware.

Average package power does not follow a single cross-benchmark pattern. Most
M5 Pro dense-model rows have power ratios below one, whereas the LiveCodeBench
ratios are $1.02$ and $1.16$. We therefore treat wall time and total energy as
the primary systems outcomes rather than attributing the cross-chip differences
to power alone.

The served GGUF checkpoints occupy $6.80$, $15.24$, and $29.55\GB$ for
Coder-3B, Coder-7B, and Coder-14B, respectively. None of the dense-model runs
uses swap. At $K=1$, measured peak process RSS is approximately $7.62$,
$15.95$, and $32.48\GB$ for the three models, and it rises with retry depth to
about $47.1\GB$ for Coder-14B at $K=8$. This increase is larger than the
computed KV-cache growth alone predicts, so the main text uses checkpoint
weights plus computed KV cache for model-memory comparisons and treats RSS as a
separate physical measurement. The roughly $47$--$48\GB$ peak at
Coder-14B $K=8$ also motivates the use of 64-GiB systems for the full
experiment.

{\bf Device-generation accuracy check.\enspace}
We verify that the physical-device serving path preserves candidate quality.
On the full 164-task HumanEval+ population, independent fp16-GGUF/llama.cpp
generations differ from the bf16/vLLM pools by at most $1.23$ percentage
points in \passone{} and $1.21$ points in \oraclek{$K$} across Coder-14B at
$K=1$, Coder-7B at $K=2$, and Coder-3B at $K=4$. The differences are small
and similar across models and retry budgets, supporting the use of
full-population pool accuracy alongside the physical-device systems
measurements.

Accuracy and systems cost are measured separately. Accuracy comes from the full
offline candidate populations using public-test selection, while wall time,
energy, and power are measured on the physical devices. The M3 Max dense-model
measurements cover the full 164, 378, 342, and 175 task populations for
HumanEval+, MBPP+, BigCodeBench, and LiveCodeBench, respectively. On M5 Pro,
HumanEval+ uses two 82-task measurements; MBPP+ uses a 190-task first
measurement and a 95-task second measurement; BigCodeBench covers 170 tasks;
and LiveCodeBench uses one 44-task shard. Table~\ref{tab:apple-main} reports
the geometric mean of the two HumanEval+ and MBPP+ measurement ratios and the
available BigCodeBench and LiveCodeBench measurements directly.

The M5 Pro LiveCodeBench measurement was run on a second M5 Pro chassis
(\texttt{Mac17,9}) rather than the \texttt{Mac17,8} system used for the other
M5 Pro measurements. The systems report the same Apple M5 Pro chip, 18 logical
CPUs, 64 GiB of RAM, and the same Metal working-set limit. Single-stream decode
rates for the 3B and 7B checkpoints differ by at most $1.3\%$ between the two
systems, so we report the result as M5 Pro while retaining the chassis
provenance.

\subsection{Conventional 4/8-bit quantization}
\label{app:composition-values}

We next ask whether the same pattern holds for more conventional 4- and 8-bit
quantization. For each model size, Table~\ref{tab:precision-direct} compares
bf16 with the measured quantized checkpoint that uses the least weight memory.
The choice is based only on checkpoint size, not benchmark accuracy.

\begin{table*}[t]
\centering
\scriptsize
\setlength{\tabcolsep}{2.9pt}
\renewcommand{\arraystretch}{1.07}

\caption{\textbf{Conventional quantization substantially reduces weight memory with relatively small changes in accuracy.}
Each row compares bf16 with the lowest-memory measured precision at the same
model size. Accuracy differences are compact minus bf16, in percentage points.
The MoE result uses a third-party AWQ4 checkpoint. Full results for every
precision are given in Tables~\ref{tab:composition} and~\ref{tab:moe-precision-fan}.}
\label{tab:precision-direct}

\resizebox{\textwidth}{!}{%
\begin{tabular}{@{}llrr rr rr rr rr r@{}}
\toprule
& & \multicolumn{2}{c}{\textbf{Memory}}
& \multicolumn{2}{c}{\textbf{MBPP+}}
& \multicolumn{2}{c}{\textbf{HumanEval+}}
& \multicolumn{2}{c}{\textbf{LiveCodeBench}}
& \multicolumn{2}{c}{\textbf{BigCodeBench}}
& \\
\cmidrule(lr){3-4}
\cmidrule(lr){5-6}
\cmidrule(lr){7-8}
\cmidrule(lr){9-10}
\cmidrule(lr){11-12}

Model size
& \shortstack{Compact\\precision}
& \shortstack{Weights (GB)\\bf16 $\to$ compact}
& \shortstack{Weight\\saving}
& $\Delta$\passone & $\Delta$\execk{50}
& $\Delta$\passone & $\Delta$\execk{50}
& $\Delta$\passone & $\Delta$\execk{50}
& $\Delta$\passone & $\Delta$\execk{50}
& \shortstack{Largest $|\Delta|$\\at $K=50$} \\
\midrule

Coder-3B
& GPTQ4
& 6.17 $\to$ 2.07 & $2.98\times$
& +0.7 & +0.0
& -1.0 & +1.7
& +0.0 & +2.3
& -1.7 & -0.3
& 2.3 \\

Coder-7B
& AWQ4
& 15.23 $\to$ 5.57 & $2.73\times$
& +2.0 & +1.3
& -2.6 & -3.1
& -0.2 & -2.3
& -1.6 & +0.1
& 3.1 \\

Coder-32B
& AWQ4
& 65.53 $\to$ 19.33 & $3.39\times$
& -0.7 & -3.0
& -0.6 & +0.9
& +0.6 & +2.3
& +0.3 & +0.1
& 3.0 \\

Qwen3-Coder MoE
& AWQ4
& 61.07 $\to$ 16.81 & $3.63\times$
& +0.4 & +0.3
& +0.2 & -0.7
& -1.8 & -2.3
& -3.4 & +0.8
& 2.3 \\

\bottomrule
\end{tabular}}
\end{table*}

The reduction in weight memory is substantial, ranging from about
$2.7\times$ to $3.6\times$, while the corresponding accuracy changes are much
smaller and vary in direction. Across these comparisons, the largest change
after 50 verified attempts is $3.1$ percentage points. Conventional
quantization therefore preserves retried accuracy fairly well on these
benchmarks. The one-bit Bonsai experiment is a more severe intervention and
produces the larger one-shot loss needed to study recovery under aggressive
compression.

Table~\ref{tab:composition} gives the complete Qwen2.5-Coder precision sweep,
including the intermediate precisions omitted above. We report both public-test
selection and CodeT because stability under an external verifier does not
necessarily imply stability when the model must generate its own tests.

\begin{table*}[!t]
\centering
\scriptsize
\setlength{\tabcolsep}{1.8pt}
\renewcommand{\arraystretch}{1.05}

\caption{\textbf{Qwen2.5-Coder accuracy across all measured precisions.}
For each model size, we report one-shot accuracy, accuracy after 50 attempts
selected with public tests, and accuracy after 50 attempts selected with
model-generated tests. The memory column reports checkpoint weight size.
LiveCodeBench CodeT uses independently generated candidate pools.}
\label{tab:composition}

\resizebox{\textwidth}{!}{%
\begin{tabular}{@{}llr rrr rrr rrr rrr@{}}
\toprule
Model
& Precision
& \shortstack{Weights\\(GB)}
& \multicolumn{3}{c}{MBPP+}
& \multicolumn{3}{c}{HumanEval+}
& \multicolumn{3}{c}{LiveCodeBench}
& \multicolumn{3}{c}{BigCodeBench} \\
\cmidrule(lr){4-6}
\cmidrule(lr){7-9}
\cmidrule(lr){10-12}
\cmidrule(l){13-15}

& & &
\passone & \execk{50} & \codetk{50} &
\passone & \execk{50} & \codetk{50} &
\passone & \execk{50} & \codetk{50} &
\passone & \execk{50} & \codetk{50} \\
\midrule

\multirow{4}{*}{Coder-3B}
& bf16       & 6.17 & 61.6\% & 77.5\% & 75.4\% & 77.4\% & 87.9\% & 84.2\% &  9.5\% & 26.3\% & 18.3\% & 41.3\% & 64.7\% & 49.7\% \\
& GPTQ 8-bit & 3.47 & 63.3\% & 78.3\% & 71.4\% & 77.7\% & 88.4\% & 82.9\% & 10.1\% & 28.6\% & 19.4\% & 40.7\% & 64.1\% & 48.6\% \\
& AWQ 4-bit  & 2.69 & 62.6\% & 78.0\% & 71.4\% & 75.9\% & 90.9\% & 84.1\% &  8.7\% & 26.3\% & 17.7\% & 39.5\% & 64.0\% & 47.7\% \\
& GPTQ 4-bit & 2.07 & 62.3\% & 77.5\% & 71.2\% & 76.4\% & 89.6\% & 83.5\% &  9.5\% & 28.6\% & 17.7\% & 39.5\% & 64.4\% & 47.6\% \\

\midrule
\multirow{4}{*}{Coder-7B}
& bf16       & 15.23 & 69.6\% & 78.6\% & 77.5\% & 82.6\% & 90.9\% & 84.1\% & 16.3\% & 33.7\% & 28.0\% & 47.9\% & 67.8\% & 56.2\% \\
& GPTQ 8-bit &  8.86 & 71.6\% & 80.2\% & 76.2\% & 81.4\% & 89.0\% & 83.5\% & 16.2\% & 34.3\% & 26.9\% & 47.8\% & 67.6\% & 55.5\% \\
& AWQ 4-bit  &  5.57 & 71.6\% & 79.9\% & 78.6\% & 80.0\% & 87.8\% & 83.5\% & 16.1\% & 31.4\% & 28.0\% & 46.3\% & 67.9\% & 53.5\% \\
& GPTQ 4-bit &  5.58 & 71.2\% & 79.9\% & 76.7\% & 80.8\% & 90.2\% & 84.1\% & 16.0\% & 31.4\% & 23.4\% & 47.1\% & 68.0\% & 55.4\% \\

\midrule
\multirow{4}{*}{Coder-32B}
& bf16       & 65.53 & 77.9\% & 82.9\% & 81.1\% & 85.1\% & 89.3\% & 85.7\% & 27.1\% & 38.3\% & 32.0\% & 56.6\% & 69.7\% & 62.7\% \\
& GPTQ 8-bit & 35.07 & 77.2\% & 81.1\% & 79.0\% & 85.1\% & 89.6\% & 86.0\% & 26.5\% & 41.1\% & 30.9\% & 56.8\% & 70.4\% & 62.7\% \\
& AWQ 4-bit  & 19.33 & 77.2\% & 79.9\% & 79.1\% & 84.5\% & 90.2\% & 86.0\% & 27.7\% & 40.6\% & 30.9\% & 56.9\% & 69.8\% & 63.0\% \\
& GPTQ 4-bit & 19.34 & 76.2\% & 80.4\% & 77.8\% & 84.4\% & 88.6\% & 86.2\% & 26.5\% & 38.9\% & 32.0\% & 56.1\% & 69.7\% & 61.2\% \\

\bottomrule
\end{tabular}}
\end{table*}

No quantized checkpoint is uniformly best across benchmarks, and small positive
or negative differences should not be interpreted as a monotone effect of
precision. LiveCodeBench CodeT is computed from independently generated
candidate and test pools, so those values can be compared across checkpoints
but are not paired with the public-test generations in the same row.

% For BigCodeBench, generated-test coverage is complete except for Coder-3B bf16
% and Coder-7B bf16, which each lack one usable test suite, and Coder-3B AWQ4,
% which lacks two. Whenever we report a CodeT improvement, the corresponding
% one-shot baseline is evaluated on exactly the same tasks.

% -----------------------------------------------------------------------------
\subsection{Candidate coverage under model scaling}
\label{app:oracle-coverage}

Choosing a smaller model can change more than the probability of success on a
single attempt. It can also change which tasks the model is able to solve at
all. The Qwen2.5-Coder size ladder lets us separate these effects. If one-shot
accuracy improves with model size but the fraction of tasks solved at least
once within 50 attempts stays nearly unchanged, then retry can recover much of
the gap. If that fraction also increases with model size, some of the advantage
of the larger model cannot be recovered by sampling alone.

\begin{figure*}[t]
\centering
{\setlength{\tabcolsep}{2pt}%
\begin{tabular}{@{}cc@{}}
\multicolumn{2}{c}{
  \includegraphics[width=\textwidth]{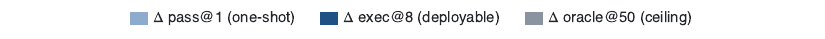}
} \\[-2pt]

\includegraphics[width=0.45\textwidth]{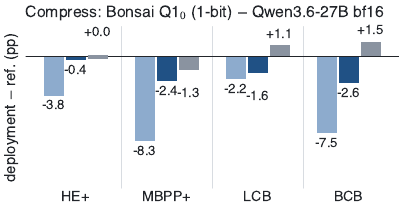} &
\includegraphics[width=0.45\textwidth]{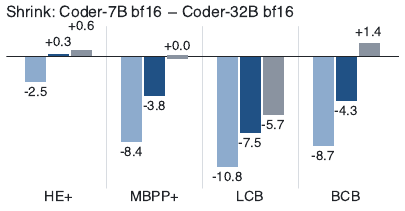}
\\[-2pt]

\small (a) Compress: fewer bits &
\small (b) Shrink: fewer weights
\end{tabular}}

\caption{\textbf{Retry recovers lost one-shot accuracy only when correct solutions remain reachable.}
Panel (a) compares one-bit Bonsai with its bf16 Qwen3.6-27B parent, and panel
(b) compares Coder-7B with Coder-32B. Bars show the change relative to the
higher-precision or larger reference for accuracy after eight verified attempts
and for the fraction of tasks solved at least once within 50 attempts.
Quantization largely preserves candidate coverage, whereas reducing model size
can lower it, most clearly on LiveCodeBench.}
\label{fig:memory-routes}
\end{figure*}

\begin{table*}[t]
\centering
\scriptsize
\setlength{\tabcolsep}{3.2pt}
\renewcommand{\arraystretch}{1.08}

\caption{\textbf{Model scale mostly improves single-attempt reliability on
MBPP+, HumanEval+, and BigCodeBench, but also expands the set of solvable tasks
on LiveCodeBench.}
The endpoint columns compare 3B with 32B. The \oraclek{50} column shows all
four native model sizes (3B / 7B / 14B / 32B), and the final column reports
their range. Accuracy differences are in percentage points. LiveCodeBench
CodeT uses independently generated candidate pools. }
\label{tab:oracle-coverage}

\resizebox{\textwidth}{!}{%
\begin{tabular}{@{}lcccccc@{}}
\toprule
Benchmark
& \shortstack{\passone\\3B $\to$ 32B}
& $\Delta$\passone
& \shortstack{\execk{50}\\3B $\to$ 32B}
& \shortstack{\codetk{50}\\3B $\to$ 32B}
& \shortstack{\oraclek{50}\\3B / 7B / 14B / 32B}
& \shortstack{\oraclek{50}\\range} \\
\midrule

MBPP+
& $61.6\% \to 77.9\%$
& +16.3
& $77.5\% \to 82.9\%$
& $75.4\% \to 81.1\%$
& $86.8/87.3/86.0/87.3\%$
& 1.3 \\

HumanEval+
& $77.4\% \to 85.1\%$
& +7.7
& $87.9\% \to 89.3\%$
& $84.2\% \to 85.7\%$
& $93.9/92.7/94.5/92.1\%$
& 2.4 \\

BigCodeBench
& $41.3\% \to 56.6\%$
& +15.4
& $64.7\% \to 69.7\%$
& $49.7\% \to 62.7\%$
& $75.6/76.9/77.0/75.5\%$
& 1.5 \\

LiveCodeBench
& $9.5\% \to 27.1\%$
& +17.6
& $26.3\% \to 38.3\%$
& $18.3\% \to 32.0\%$
& $30.3/37.1/43.4/42.9\%$
& 13.1 \\

\bottomrule
\end{tabular}}
\end{table*}

Table~\ref{tab:oracle-coverage} shows a striking separation between
one-shot accuracy and what the models can solve after repeated attempts on
the first three benchmarks. From 3B to
32B, one-shot accuracy rises by $16.3$, $7.7$, and $15.4$ percentage points
on MBPP+, HumanEval+, and BigCodeBench, respectively. Yet the fraction of tasks
solved at least once within 50 attempts varies by only $1.3$, $2.4$, and
$1.5$ points across the entire model-size ladder. On these benchmarks, much of
the advantage of the larger model comes from solving the same tasks more
reliably on each attempt rather than from being able to solve many additional
tasks.

LiveCodeBench is different. One-shot accuracy rises by $17.6$ percentage
points from 3B to 32B, and the fraction of tasks solved at least once within
50 attempts also changes substantially, by $13.1$ points. For Coder-7B,
public-test selection reaches $33.7\%$ at $K=50$, already close to the
$37.1\%$ of tasks for which the model generates at least one correct solution.
The remaining gap to the larger models therefore cannot be closed mainly by a
better selector: the smaller model often never generates a correct solution.

Quantizing a larger model provides another option when memory permits. On
MBPP+, for example, Coder-32B AWQ4 and Coder-7B AWQ4 both reach $79.9\%$ after
50 verified attempts, while their checkpoints occupy $19.33\GB$ and
$5.57\GB$, respectively. On HumanEval+ and LiveCodeBench, however, the larger
checkpoint retains an accuracy advantage. The useful deployment choice
therefore depends on whether additional model scale mainly improves reliability
or actually expands the set of tasks the model can solve.
% -----------------------------------------------------------------------------
\subsection{Precision replication in a sparse mixture-of-experts model}
\label{app:moe-fan}

We also test whether the precision results carry over to a sparse
mixture-of-experts model. Qwen3-Coder-30B-A3B has 30.5B total parameters but
activates only 3.3B parameters per token, and is available in precisions ranging
from bf16 to 4-bit. This provides a useful contrast with the dense
Qwen2.5-Coder models.

\begin{table*}[!t]
\centering
\scriptsize
\setlength{\tabcolsep}{3.2pt}
\renewcommand{\arraystretch}{1.05}

\caption{\textbf{Qwen3-Coder-30B-A3B remains relatively stable across precision levels.}
The model has 30.5B total parameters and 3.3B active parameters per token.
The final two rows report the range across precisions for one-shot accuracy and
accuracy after 50 verified attempts; brackets give paired 95\% bootstrap
intervals for the latter.}
\label{tab:moe-precision-fan}

\resizebox{\textwidth}{!}{%
\begin{tabular}{@{}lr rr rr rr rr@{}}
\toprule
Precision & \shortstack{Weights\\(GB)}
& \multicolumn{2}{c}{MBPP+}
& \multicolumn{2}{c}{HumanEval+}
& \multicolumn{2}{c}{LiveCodeBench}
& \multicolumn{2}{c}{BigCodeBench} \\
\cmidrule(lr){3-4}
\cmidrule(lr){5-6}
\cmidrule(lr){7-8}
\cmidrule(l){9-10}

& & \passone & \execk{50}
& \passone & \execk{50}
& \passone & \execk{50}
& \passone & \execk{50} \\
\midrule

bf16
& 61.07
& 77.2\% & 79.3\%
& 88.5\% & 90.9\%
& 32.7\% & 43.4\%
& 57.0\% & 63.3\% \\

FP8
& 31.18
& 77.1\% & 79.4\%
& 88.2\% & 91.5\%
& 31.6\% & 45.1\%
& 57.8\% & 63.5\% \\

GPTQ 8-bit
& 32.00
& 77.4\% & 79.1\%
& 88.2\% & 90.9\%
& 32.4\% & 45.1\%
& 54.7\% & 62.6\% \\

AWQ 4-bit
& 16.81
& 77.6\% & 79.6\%
& 88.7\% & 90.2\%
& 30.9\% & 41.1\%
& 53.6\% & 64.1\% \\

GPTQ 4-bit
& 16.93
& 76.8\% & 78.8\%
& 87.8\% & 90.2\%
& 31.5\% & 44.0\%
& 52.6\% & 62.1\% \\

\addlinespace[2pt]
\multicolumn{2}{@{}l}{\passone{} range}
& \multicolumn{2}{c}{0.8}
& \multicolumn{2}{c}{0.9}
& \multicolumn{2}{c}{1.8}
& \multicolumn{2}{c}{5.2} \\

\multicolumn{2}{@{}l}{\execk{50} range [95\% CI]}
& \multicolumn{2}{c}{$0.8\;[-0.5,+2.1]$}
& \multicolumn{2}{c}{$1.2\;[+0.0,+3.0]$}
& \multicolumn{2}{c}{$\mathbf{4.0\;[+0.6,+8.0]}$}
& \multicolumn{2}{c}{$2.1\;[-1.2,+5.3]$} \\

\bottomrule
\end{tabular}}
\end{table*}

Table~\ref{tab:moe-precision-fan} shows that the same pattern largely extends
to the sparse MoE model. Despite a more than threefold difference in weight
memory between bf16 and 4-bit checkpoints, accuracy changes relatively little
on most benchmarks. After 50 verified attempts, the full precision range is
only $0.8$ percentage points on MBPP+, $1.2$ on HumanEval+, and $2.1$ on
BigCodeBench. LiveCodeBench is less stable, with a $4.0$-point range and a
paired interval that excludes zero. BigCodeBench shows the opposite pattern:
its $5.2$-point one-shot spread shrinks to $2.1$ points after 50 attempts.
Thus, retry often reduces sensitivity to precision, but not uniformly across
workloads.

The sparse architecture also changes the relationship between memory and
serving cost. Across the four Qwen3-Coder workloads, throughput remains between
18.6k and 19.5k tokens/s. On LiveCodeBench, the MoE model reaches
19,520 tokens/s, compared with 7,852 tokens/s for dense Coder-32B under the
matched protocol, while using $2.83\times$ less energy per token. The expert
weights still occupy memory, but only a small fraction of the model is active
for each token.

We exclude DeepSeek-Coder-6.7B from the quantitative precision comparison
because its MBPP+ generations contain 2,006 structurally malformed,
concatenated solutions. This is a generation-format failure rather than an
effect of numerical precision.

% -----------------------------------------------------------------------------
\subsection{BigCodeBench verification limits}
\label{app:bigcodebench-full}

BigCodeBench makes the gap between finding a correct solution and identifying
it especially clear. Table~\ref{tab:bigcodebench-full} compares public-test
selection, candidate coverage, and CodeT across the full Qwen2.5-Coder
precision sweep. %Public-test results use the same $1{,}034$ tasks for every
%checkpoint, while CodeT uses the subset of tasks for which the corresponding
%generated-test results are available.

\begin{table*}[!t]
\centering
\scriptsize
\setlength{\tabcolsep}{3.3pt}
\renewcommand{\arraystretch}{1.05}

\caption{\textbf{Verification, rather than candidate generation, remains a major bottleneck on BigCodeBench.}}
\label{tab:bigcodebench-full}

\resizebox{\textwidth}{!}{%
\begin{tabular}{@{}llrrrrrr@{}}
\toprule
Model
& Precision
& \execk{50}
& \oraclek{50}
& \shortstack{Public-test\\gap}
& \codetk{50}
& \shortstack{CodeT\\gap}
& \shortstack{Tasks with\\usable tests} \\
\midrule

\multirow{4}{*}{Coder-3B}
& bf16       & 64.7\% & 75.6\% & 10.9 & 49.7\% & 15.0 & 99.9\% \\
& GPTQ 8-bit & 64.1\% & 74.5\% & 10.3 & 48.6\% & 15.5 & 100.0\% \\
& AWQ 4-bit  & 64.0\% & 75.1\% & 11.1 & 47.7\% & 16.3 & 99.8\% \\
& GPTQ 4-bit & 64.4\% & 75.5\% & 11.1 & 47.6\% & 16.8 & 100.0\% \\

\midrule
\multirow{4}{*}{Coder-7B}
& bf16       & 67.8\% & 76.9\% & 9.1 & 56.2\% & 11.6 & 99.9\% \\
& GPTQ 8-bit & 67.6\% & 76.7\% & 9.1 & 55.5\% & 12.1 & 100.0\% \\
& AWQ 4-bit  & 67.9\% & 76.4\% & 8.5 & 53.5\% & 14.4 & 100.0\% \\
& GPTQ 4-bit & 68.0\% & 77.0\% & 9.0 & 55.4\% & 12.6 & 100.0\% \\

\midrule
Coder-14B
& bf16       & 69.9\% & 77.0\% & 7.1 & 60.8\% & 9.1 & 100.0\% \\

\midrule
\multirow{4}{*}{Coder-32B}
& bf16       & 69.7\% & 75.5\% & 5.8 & 62.7\% & 7.0 & 100.0\% \\
& GPTQ 8-bit & 70.4\% & 75.0\% & 4.6 & 62.7\% & 7.7 & 100.0\% \\
& AWQ 4-bit  & 69.8\% & 75.0\% & 5.2 & 63.0\% & 6.8 & 100.0\% \\
& GPTQ 4-bit & 69.7\% & 75.6\% & 5.9 & 61.2\% & 8.5 & 100.0\% \\

\bottomrule
\end{tabular}}
\end{table*}

Table~\ref{tab:bigcodebench-full} shows that, for the native bf16 models, the
gap between public-test selection and the best candidate available after 50
attempts shrinks with model size, from $10.9$ percentage points at 3B to
$5.8$ points at 32B. At the same time, the fraction of tasks with at least one
correct candidate changes very little, remaining within a $1.5$-point range
across the model-size ladder. Larger models therefore make the available
correct candidates easier to identify rather than greatly expanding the set of
tasks solved within 50 attempts.

The same table also shows that model-generated tests leave an even larger gap
for many checkpoints. Test-generation failures are rare, so the deeper
$K=500$ analysis in Section~\ref{sec:ceiling} points to the main limitation. On many tasks, correct and
incorrect programs produce the same outcomes on the generated tests, leaving
too little information for the selector to tell them apart.

% -----------------------------------------------------------------------------
\subsection{Alternative compact deployment strategies}
\label{app:compact-alternatives}

A practical deployment need not use only a smaller or quantized Qwen model.
It may instead use a distilled, pruned, reasoning-distilled, or entirely
different model family. We therefore compare several compact checkpoints after
50 verified attempts. Because these models differ in architecture, training
data, and training objective, the comparison is descriptive rather than a
controlled comparison of compression methods.

\begin{figure*}[t]
\centering
\includegraphics[width=\textwidth]{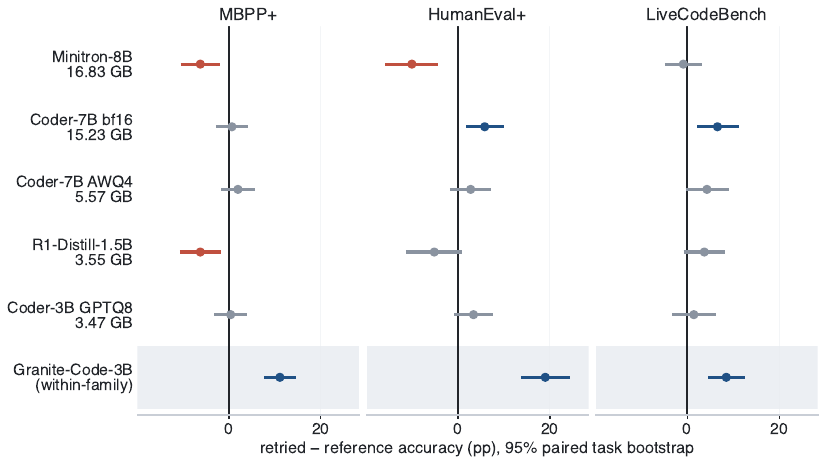}
\caption{\textbf{Verified retry has different effects across compact checkpoints.}
Points show retried accuracy minus one-shot reference accuracy, with 95\%
paired task-bootstrap intervals. Each comparison uses its common task set, so
the plotted differences need not exactly match subtraction of the rounded
full-population values in Table~\ref{tab:deployment-strategies}. Blue intervals
are above zero, red intervals are below zero, and gray intervals cross zero.
Granite-Code uses Granite-Code-34B as its reference rather than
Qwen2.5-Coder-32B, so its effect sizes are not directly comparable with the
other rows.}
\label{fig:strategy-forest}
\end{figure*}

Figure~\ref{fig:strategy-forest} shows the paired accuracy difference between
each retried checkpoint and its one-shot reference. Intervals that lie entirely
above or below zero indicate a detectable advantage or deficit, respectively;
intervals that cross zero are consistent with the reference at the 95\% level.
Granite-Code uses its own within-family 34B reference and is shown separately.

\begin{table*}[!t]
\centering
\scriptsize
\setlength{\tabcolsep}{2.5pt}
\renewcommand{\arraystretch}{1.05}

\caption{\textbf{Verified retry across several compact deployment strategies.}
Each row compares a compact checkpoint after 50 verified attempts with the
one-shot reference shown in the third column. Accuracy is retried/reference,
and $\Delta$ is the paired accuracy difference where confidence intervals are
available. Granite-Code uses Granite-Code-34B as its within-family reference.}
\label{tab:deployment-strategies}

\resizebox{\textwidth}{!}{%
\begin{tabular}{@{}lllcrr cc cc cc@{}}
\toprule
Strategy
& Retried checkpoint
& One-shot reference
& Selection
& \shortstack{Retried weights\\(GB)}
& \shortstack{Weight\\saving}
& \multicolumn{2}{c}{MBPP+}
& \multicolumn{2}{c}{HumanEval+}
& \multicolumn{2}{c}{LiveCodeBench} \\
\cmidrule(lr){7-8}
\cmidrule(lr){9-10}
\cmidrule(l){11-12}

& & & & & &
Acc. & $\Delta$ &
Acc. & $\Delta$ &
Acc. & $\Delta$ \\
\midrule

Prune + distill
& Minitron-8B
& Coder-32B bf16
& \execk{50}
& 16.83
& $3.9\times$
& 71.7/77.9\%
& $\mathbf{-6.2\;[-10.5,-2.0]}$
& 75.0/85.1\%
& $\mathbf{-10.0\;[-15.9,-4.4]}$
& 26.3/27.1\%
& $-1.1\;[-5.1,+3.0]$ \\

Native bf16
& Coder-7B
& Coder-32B bf16
& \execk{50}
& 15.23
& $4.3\times$
& 78.6/77.9\%
& $+0.7\;[-2.8,+4.2]$
& 90.9/85.1\%
& $\mathbf{+5.8\;[+1.8,+10.0]}$
& 33.7/27.1\%
& $\mathbf{+6.3\;[+1.8,+11.1]}$ \\

Quantization
& Coder-7B AWQ4
& Coder-32B bf16
& \execk{50}
& 5.57
& $11.8\times$
& 79.9/77.9\%
& $+2.0\;[-1.7,+5.7]$
& 87.8/85.1\%
& $+2.8\;[-1.8,+7.3]$
& 31.4/27.1\%
& $+4.0\;[-0.6,+8.8]$ \\

Reasoning distill
& R1-Distill-1.5B
& Coder-32B bf16
& \execk{50}
& 3.55
& $18.5\times$
& 71.7/77.9\%
& $\mathbf{-6.2\;[-10.7,-1.7]}$
& 79.9/85.1\%
& $-5.2\;[-11.4,+0.9]$
& 30.9/27.1\%
& $+3.5\;[-1.1,+8.0]$ \\

Quantization
& Coder-3B GPTQ8
& Coder-32B bf16
& \execk{50}
& 3.47
& $18.9\times$
& 78.3/77.9\%
& $+0.4\;[-3.2,+3.9]$
& 88.4/85.1\%
& $+3.4\;[-0.9,+7.7]$
& 28.6/27.1\%
& $+1.2\;[-3.6,+5.9]$ \\

\midrule
External family
& Granite-Code-3B
& Granite-Code-34B
& \execk{50}
& --
& $9.7\times$
& 66.7/55.6\%
& +11.1
& 72.0/53.0\%
& +19.0
& 17.7/9.2\%
& +8.5 \\

\bottomrule
\end{tabular}}
\end{table*}

Figure~\ref{fig:strategy-forest} and
Table~\ref{tab:deployment-strategies} show that the outcome depends strongly
on the compact checkpoint. Native Coder-7B exceeds the Coder-32B one-shot
reference on HumanEval+ and LiveCodeBench with confidence intervals above zero.
Coder-7B AWQ4 and Coder-3B GPTQ8 are statistically consistent with the 32B
reference on all three benchmarks. Minitron and R1-Distill, in contrast, retain
a clear deficit on at least one benchmark. Granite-Code provides a separate
within-family example: Granite-Code-3B with 50 verified attempts exceeds the
Granite-Code-34B one-shot reference on all three benchmarks while using
$9.7\times$ fewer weight bytes.

The ordering can also change when the verifier changes. Minitron, for example,
is weaker under public-test selection on MBPP+ and HumanEval+ but produces some
of the largest improvements from CodeT in our experiments. A model's ability
to generate good candidate solutions and its ability to generate useful
verification evidence are therefore separate properties.

% -----------------------------------------------------------------------------
\subsection{Practical retry budgets}
\label{app:operating-points}

We use $K=50$ to study how much capability remains reachable after repeated
sampling, but such a large retry budget is not necessarily useful in practice.
A more relevant question is how quickly the accuracy gains arrive. On MBPP+,
Table~\ref{tab:retry-operating-points} compares the fraction of the eventual
$K=50$ gain obtained at smaller values of $K$ with the corresponding increase
in wall-clock time.

\begin{table*}[!t]
\centering
\small
\setlength{\tabcolsep}{5.5pt}
\renewcommand{\arraystretch}{1.05}

\caption{\textbf{Most of the benefit of retry arrives at moderate budgets.}
For each model, ``Gain'' is the fraction of its improvement from $K=1$ to
$K=50$ that has already been obtained at the displayed $K$. Wall time is
reported relative to the same model at $K=1$.}
\label{tab:retry-operating-points}

\begin{tabular}{@{}r cc cc cc@{}}
\toprule
& \multicolumn{2}{c}{Coder-3B}
& \multicolumn{2}{c}{Coder-7B}
& \multicolumn{2}{c}{Coder-32B} \\
\cmidrule(lr){2-3}
\cmidrule(lr){4-5}
\cmidrule(l){6-7}

$K$
& Gain (\%)
& \shortstack{Wall-time\\multiplier}
& Gain (\%)
& \shortstack{Wall-time\\multiplier}
& Gain (\%)
& \shortstack{Wall-time\\multiplier} \\
\midrule

1  & 0   & 1.00  & 0   & 1.00 & 0   & 1.00 \\
2  & 40  & 1.45  & 56  & 1.14 & 33  & 1.12 \\
\bfseries 4
   & \bfseries 63 & \bfseries 1.44
   & \bfseries 83 & \bfseries 1.35
   & \bfseries 55 & \bfseries 1.29 \\
\bfseries 8
   & \bfseries 80 & \bfseries 1.95
   & \bfseries 93 & \bfseries 1.73
   & \bfseries 80 & \bfseries 1.72 \\
16 & 91  & 4.78  & 99  & 3.13 & 90  & 2.68 \\
32 & 97  & 9.13  & 99  & 5.80 & 100 & 4.80 \\
50 & 100 & 13.55 & 100 & 8.91 & 100 & 7.28 \\

\bottomrule
\end{tabular}
\end{table*}

Table~\ref{tab:retry-operating-points} shows that the useful gains arrive well
before $K=50$. For Coder-7B, four attempts recover $83\%$ of the eventual
improvement at only $1.35\times$ the one-shot wall time, while eight attempts
recover $93\%$ at $1.73\times$ the wall time. The same broad pattern holds for
3B and 32B. Beyond about eight attempts, additional accuracy becomes small
relative to the increase in serving time. We therefore use large retry budgets
mainly to measure reachable capability, while the deployment comparisons focus
on smaller values of $K$.

% -----------------------------------------------------------------------------
\subsection{Public-test fallback behavior}
\label{app:fallback}

The public-test selector chooses among candidates that pass all visible tests.
If none passes, it returns the candidate with the highest mean token log-probability. Thus, the fallback is used only when the visible tests provide
no passing candidate.

\begin{table*}[t]
\centering
\scriptsize
\setlength{\tabcolsep}{5.0pt}
\renewcommand{\arraystretch}{1.06}

\caption{\textbf{Retry sharply reduces how often the selector must fall back.}
The left block varies the retry budget for Coder-7B; values below $K=50$ are
averaged over 120 candidate subsets and $K=50$ is exact. The right block
compares native model sizes at $K=50$.}
\label{tab:fallback-k}

\resizebox{\textwidth}{!}{%
\begin{tabular}{@{}lrrr rrrr@{}}
\toprule
& \multicolumn{3}{c}{\textbf{Coder-7B across retry budget}}
& \multicolumn{4}{c}{\textbf{Model size at $K=50$}} \\
\cmidrule(lr){2-4}
\cmidrule(l){5-8}

Benchmark & $K=1$ & $K=8$ & $K=50$ & 3B & 7B & 14B & 32B \\
\midrule
MBPP+         & 18.2\% & 6.0\%  & 4.2\%  & 5.0\%  & 4.2\%  & 4.0\%  & 4.2\% \\
HumanEval+    & 12.4\% & 5.8\%  & 4.3\%  & 3.7\%  & 4.3\%  & 3.7\%  & 6.1\% \\
LiveCodeBench & 77.8\% & 55.4\% & 38.9\% & 48.6\% & 38.9\% & 30.9\% & 34.3\% \\
BigCodeBench  & 40.1\% & 20.8\% & 13.9\% & 14.1\% & 13.9\% & 14.5\% & 15.1\% \\
\bottomrule
\end{tabular}}
\end{table*}

Table~\ref{tab:fallback-k} shows that additional attempts substantially reduce
fallback. For Coder-7B on LiveCodeBench, for example, the fallback rate falls
from $77.8\%$ with one attempt to $38.9\%$ with 50. By contrast, at $K=50$
model size has relatively little effect on fallback for MBPP+, HumanEval+, and
BigCodeBench. On LiveCodeBench, public-test selection already recovers most of
the correct candidates produced by Coder-7B, so the remaining advantage of
larger models comes mainly from generating solutions that the 7B model does
not produce.

% -----------------------------------------------------------------------------
\subsection{Serial reasoning versus parallel retry}
\label{app:thinking}

Additional test-time compute can be spent in two different ways: generating
several independent attempts or producing a longer reasoning trace before
returning one answer. Qwen3-30B-A3B provides Instruct and Thinking variants
with the same architecture and weight footprint, allowing us to compare these
two strategies without changing model size. This experiment uses a
$16{,}384$-token generation limit and is therefore reported separately from
the shorter-generation systems sweeps in the main text.

\begin{table*}[!t]
\centering
\scriptsize
\setlength{\tabcolsep}{3.0pt}
\renewcommand{\arraystretch}{1.05}

\caption{\textbf{Longer reasoning and parallel retry have different advantages
across benchmarks.}
The Instruct and Thinking variants use the same Qwen3-30B-A3B architecture and
weight footprint. Accuracy uses public-test selection on the full benchmark
sets; wall time and energy come from matched 40-prompt B200 workloads and
should be compared only within a benchmark. Bold rows highlight eight
non-reasoning attempts versus one reasoning attempt.}
\label{tab:thinkmode}

\resizebox{\textwidth}{!}{%
\begin{tabular}{@{}l r rrr rrr rrr@{}}
\toprule
& & \multicolumn{3}{c}{\textbf{MBPP+}}
& \multicolumn{3}{c}{\textbf{HumanEval+}}
& \multicolumn{3}{c}{\textbf{LiveCodeBench}} \\
\cmidrule(lr){3-5}
\cmidrule(lr){6-8}
\cmidrule(l){9-11}

Mode & $K$
& Acc. & Wall (s) & J/task
& Acc. & Wall (s) & J/task
& Acc. & Wall (s) & J/task \\
\midrule

\multirow{3}{*}{Instruct}
& 1
& 77.9\% & 16.5  & 243
& 87.9\% & 52.8  & 700
& 43.8\% & 114.5 & 1{,}707 \\

& \bfseries 8
& \bfseries 79.3\% & \bfseries 53.3 & \bfseries 866
& \bfseries 89.9\% & \bfseries 137.2 & \bfseries 2{,}301
& \bfseries 53.0\% & \bfseries 199.4 & \bfseries 4{,}317 \\

& 50
& 79.6\% & 130.2 & 2{,}454
& 90.2\% & 257.5 & 5{,}099
& 52.6\% & 467.6 & 11{,}156 \\

\midrule
\multirow{3}{*}{Thinking}
& \bfseries 1
& \bfseries 79.5\% & \bfseries 124.2 & \bfseries 2{,}008
& \bfseries 91.2\% & \bfseries 83.8 & \bfseries 1{,}389
& \bfseries 51.7\% & \bfseries 186.0 & \bfseries 3{,}899 \\

& 8
& 80.9\% & 218.3 & 4{,}899
& 92.9\% & 193.0 & 3{,}972
& 62.2\% & 578.2 & 13{,}670 \\

& 50
& 81.0\% & 748.1 & 17{,}948
& 94.5\% & 685.2 & 15{,}684
& 64.6\% & 3{,}020.2 & 74{,}190 \\

\bottomrule
\end{tabular}}
\end{table*}

Table~\ref{tab:thinkmode} shows that neither strategy is consistently better.
On MBPP+, eight non-reasoning attempts come within $0.2$ percentage points of
one reasoning attempt while using about $2.3\times$ less wall time and energy.
On HumanEval+, one reasoning attempt is $1.3$ points more accurate than eight
non-reasoning attempts, while also running $1.64\times$ faster and using less
energy. LiveCodeBench falls between these cases: eight non-reasoning attempts
are $1.3$ points more accurate, with similar wall time and energy.

Deeper sampling also affects the two modes differently. At $K=50$, the
Thinking model improves accuracy by about $1.3$, $4.3$, and $12.0$ percentage
points over the Instruct model on MBPP+, HumanEval+, and LiveCodeBench,
respectively. On LiveCodeBench, the gap between selected accuracy and the
fraction of tasks with a correct candidate shrinks from $12.5$ to $3.4$
percentage points, suggesting that reasoning changes not only which solutions
are generated but also how effectively the visible tests identify them. These
gains are expensive, however: at $K=50$, Thinking uses about $6.5\times$ the
wall time and $6.7\times$ the energy of the non-reasoning model.

% -----------------------------------------------------------------------------
\subsection{Additional scale checks}
\label{app:selector-values}

Our main controlled comparisons focus on the 3B, 7B, and 32B Qwen2.5-Coder
models. Table~\ref{tab:offladder-compact} adds the native 1.5B and 14B
checkpoints, together with a separate 72B Qwen2.5 model, to show how the
intermediate and larger scales behave on MBPP+.

\begin{table}[!t]
\centering
\scriptsize
\setlength{\tabcolsep}{3.5pt}

\caption{\textbf{Additional model sizes on MBPP+ at $K=50$.}
All four accuracy measures in each row are computed from the same candidate
pool. The 1.5B and 14B checkpoints belong to the native Qwen2.5-Coder scale
ladder. Qwen2.5-72B is a separate general model and is included only as an
additional accuracy comparison.}
\label{tab:offladder-compact}

\resizebox{\columnwidth}{!}{%
\begin{tabular}{@{}llrrrr@{}}
\toprule
Model & Type & \passone & \codetk{50} & \execk{50} & \oraclek{50} \\
\midrule
Coder-1.5B  & native Coder & 57.3\% & 68.3\% & 75.4\% & 84.7\% \\
Coder-14B   & native Coder & 73.2\% & 79.4\% & 80.4\% & 86.0\% \\
Qwen2.5-72B & general model & 78.2\% & 78.7\% & 80.8\% & 84.4\% \\
\bottomrule
\end{tabular}}
\end{table}

Table~\ref{tab:offladder-compact} also illustrates why model size alone does
not determine the benefit of repeated sampling. The 72B model has the highest
one-shot accuracy, but its \oraclek{50} is below that of the 14B Coder model.
We therefore treat the 72B result as an additional accuracy check rather than
as part of the controlled scaling comparison.

% -----------------------------------------------------------------------------
\subsection{Energy scaling under parallel retry}
\label{app:energy}

\begin{figure*}[t]
\centering
{\setlength{\tabcolsep}{2pt}%
\begin{tabular}{cccc}
\multicolumn{4}{c}{
  \includegraphics[width=\textwidth]{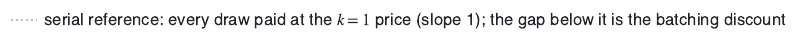}
} \\[-4pt]
\includegraphics[width=0.235\textwidth]{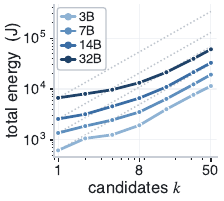} &
\includegraphics[width=0.235\textwidth]{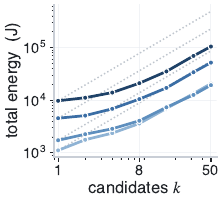} &
\includegraphics[width=0.235\textwidth]{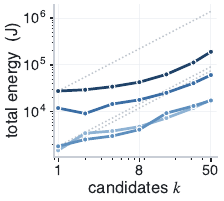} &
\includegraphics[width=0.235\textwidth]{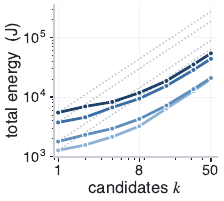} \\[-2pt]
\small (a) MBPP+ &
\small (b) HumanEval+ &
\small (c) LiveCodeBench &
\small (d) BigCodeBench
\end{tabular}}

\caption{\textbf{Parallel decoding reduces the energy cost of additional attempts.}
Solid curves show measured B200 energy for the native Qwen2.5-Coder models.
Dotted lines show the energy that would be used if every attempt cost as much
as a single attempt. Absolute energy should not be compared across panels
because the benchmarks use different workloads and generation limits.}
\label{fig:energy}
\end{figure*}

Generating $K$ candidates does not require $K$ times the energy of generating
one. Parallel decoding can make better use of the GPU, so the additional
energy grows more slowly than the number of attempts. Figure~\ref{fig:energy}
shows this effect directly.

Across the native model sizes, increasing the retry budget from one to 50
attempts raises measured energy by $6.9$--$18.6\times$, far less than the
$50\times$ increase that would result from running each attempt independently.
Parallelism therefore provides a substantial energy discount, but deeper retry
is still expensive: beyond moderate values of $K$, energy continues to grow
while the accuracy gains become progressively smaller.

The energy measurements include idle board power. We therefore compare energy
within each benchmark rather than across benchmarks, since the workloads use
different numbers of prompts and different generation limits. One
LiveCodeBench Coder-14B run uses slightly more energy at $K=1$ than at $K=2$,
most likely because of fixed startup or runtime overhead. We retain the
measurement but exclude that configuration from summaries that require a
one-shot energy baseline.

\begin{figure*}[t]
\centering
{\setlength{\tabcolsep}{2pt}%
\begin{tabular}{@{}cccc@{}}
\multicolumn{4}{c}{
  \includegraphics[width=\textwidth]{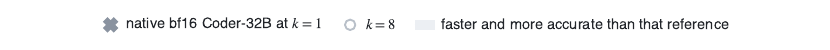}
} \\[-4pt]
\includegraphics[width=0.235\textwidth]{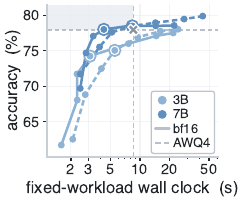} &
\includegraphics[width=0.235\textwidth]{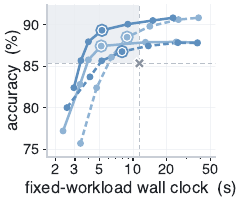} &
\includegraphics[width=0.235\textwidth]{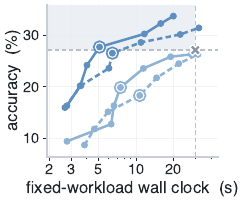} &
\includegraphics[width=0.235\textwidth]{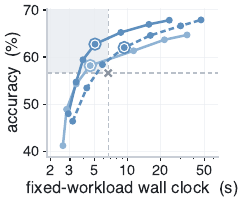} \\[-2pt]
\small (a) MBPP+ &
\small (b) HumanEval+ &
\small (c) LiveCodeBench &
\small (d) BigCodeBench
\end{tabular}}

\caption{\textbf{Quantization changes serving time as well as memory use.}
Each quantized checkpoint is compared with the bf16 model of the same size.}
\label{fig:accuracy-time-awq}
\end{figure*}

Quantization introduces a separate systems tradeoff. It substantially reduces
the memory required to store a model, but the low-bit kernels are not
necessarily faster when many sequences are decoded in parallel.
Figure~\ref{fig:accuracy-time-awq} and Table~\ref{tab:awq-batch-cost} show this
effect for Coder-7B.

\begin{table}[t]
\centering
\scriptsize
\setlength{\tabcolsep}{5.0pt}
\renewcommand{\arraystretch}{1.05}

\caption{\textbf{AWQ4 does not consistently speed up wide-batch decoding.}
Entries are Coder-7B wall-clock speedups $T_{\rm bf16}/T_{\rm AWQ4}$ on B200,
so values above $1\times$ favor AWQ4.}
\label{tab:awq-batch-cost}

\begin{tabular}{@{}lccc@{}}
\toprule
Benchmark & $K=1$ & $K=8$ & $K=50$ \\
\midrule
MBPP+         & $1.00\times$ & $0.52\times$ & $0.51\times$ \\
HumanEval+    & $1.15\times$ & $0.66\times$ & $0.60\times$ \\
LiveCodeBench & $1.04\times$ & $0.79\times$ & $0.62\times$ \\
BigCodeBench  & $0.92\times$ & $0.54\times$ & $0.52\times$ \\
\bottomrule
\end{tabular}
\end{table}

At $K=1$, Coder-7B AWQ4 is roughly as fast as bf16, with wall-clock ratios
between $0.92\times$ and $1.15\times$. At larger retry budgets, however, the
bf16 model is consistently faster: by $K=50$, the AWQ4 speedups fall to
$0.51$--$0.62\times$. Quantization therefore provides a clear memory benefit
without guaranteeing a runtime benefit, especially under wide parallel
decoding.

This distinction also helps explain why memory use and serving cost should be
reported separately. Quantization reduces the number of bits stored per
parameter, while sparse models can reduce the amount of computation performed
per token without proportionally reducing their stored weights. The resulting
effects on memory, throughput, and energy are related but not interchangeable.
% =============================================================================
\section{When the model must write its own verifier}
\label{app:verification-evidence}
% =============================================================================

Public tests provide a fixed external signal for selecting among candidate
solutions. When such tests are unavailable, the model must generate its own
tests, creating two additional ways verification can fail. The generated tests
may not distinguish correct from incorrect programs, or the selection rule may
fail to use the distinctions that are present. We study these failures
separately from the more basic case in which no correct candidate is generated.

% -----------------------------------------------------------------------------
\subsection{Aggregation with model-generated tests}
\label{app:codet}

CodeT uses patterns of agreement between candidates and generated tests rather
than simply choosing the candidate that passes the most tests. To measure the
value of this aggregation step, Table~\ref{tab:codet-delta} compares CodeT with
a simple pass-count rule while holding the candidate set and generated tests
fixed.

\begin{table}[!t]
\centering
\small

\caption{\textbf{CodeT aggregation helps most for smaller models on MBPP+.}
Values are the accuracy difference between CodeT and simple test-pass counting
at $K=50$, with paired 95\% bootstrap intervals.}
\label{tab:codet-delta}

\begin{tabular}{@{}lrr@{}}
\toprule
Model & $\Delta$ accuracy (pp) & 95\% CI (pp) \\
\midrule
Coder-1.5B & +6.08 & [ 2.65,  9.79] \\
Coder-3B   & +3.70 & [ 1.06,  6.61] \\
Coder-7B   & +3.44 & [ 1.06,  5.82] \\
Coder-14B  & +1.32 & [-0.79,  3.44] \\
Coder-32B  & +0.79 & [-1.32,  2.91] \\
\bottomrule
\end{tabular}
\end{table}

Table~\ref{tab:codet-delta} shows that CodeT provides a clear benefit for the
smaller models on MBPP+. Relative to simple pass counting, it improves accuracy
by $6.08$ percentage points at 1.5B, $3.70$ points at 3B, and $3.44$ points at
7B, with all three confidence intervals above zero. The difference becomes
small at 14B and 32B. This pattern is not universal: on HumanEval+, all of the
corresponding intervals include zero, and on LiveCodeBench some point estimates
favor simple counting. CodeT can therefore improve how generated tests are
used, but it does not remove the more fundamental limitations of
self-verification.

% -----------------------------------------------------------------------------
\subsection{Where model-generated verification fails}
\label{sec:ceiling}

\begin{figure*}[t]
\centering
{\setlength{\tabcolsep}{2pt}%
\begin{tabular}{cccc}
\multicolumn{4}{c}{
  \includegraphics[width=\textwidth]{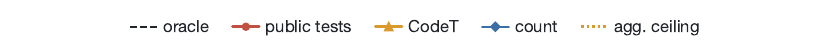}
} \\[-4pt]
\includegraphics[width=0.235\textwidth]{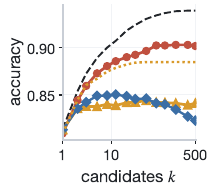} &
\includegraphics[width=0.235\textwidth]{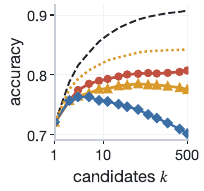} &
\includegraphics[width=0.235\textwidth]{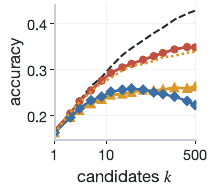} &
\includegraphics[width=0.235\textwidth]{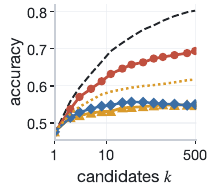} \\[-2pt]
\small (a) HumanEval+ &
\small (b) MBPP+ &
\small (c) LiveCodeBench &
\small (d) BigCodeBench
\end{tabular}}

\caption{\textbf{Deep retry separates failures of selection, generated tests,
and candidate generation.}
Curves use Qwen2.5-Coder-7B candidate pools through $K=500$. Solid curves show
public-test selection, CodeT, and simple test-pass counting. The remaining
curves show the best accuracy possible from the generated-test outcome patterns
and the fraction of tasks with at least one correct candidate. BigCodeBench has
the largest gap between these two upper bounds, showing that its generated
tests often fail to distinguish correct from incorrect programs.}
\label{fig:ceiling}
\end{figure*}

A failed selection can have three different causes. First, the selection rule
may fail to use information already present in the test outcomes. Second, the
generated tests themselves may fail to distinguish correct from incorrect
programs. Third, the candidate set may contain no correct solution at all.
Figure~\ref{fig:ceiling} separates these effects using Qwen2.5-Coder-7B with
up to 500 candidates per task.

The first gap comes from \emph{selection}. At $K=500$, CodeT reaches
$77.5\%$, $84.1\%$, $26.3\%$, and $54.7\%$ on MBPP+, HumanEval+,
LiveCodeBench, and BigCodeBench, while the best possible selection based on the
observed test-outcome patterns reaches $84.2\%$, $88.5\%$, $34.0\%$, and
$61.7\%$. The corresponding gaps of $6.7$, $4.3$, $7.7$, and $7.0$
percentage points measure how much information is present in the generated
tests but not fully extracted by CodeT.

The second gap comes from the \emph{tests themselves}. Even with the best
possible use of the observed test outcomes, accuracy remains $6.5$, $5.4$,
$8.9$, and $18.6$ percentage points below the fraction of tasks that contain a
correct candidate. BigCodeBench is the clearest example: correct and incorrect
programs often behave identically on the generated tests, so no selection rule
using only those tests can reliably tell them apart. Better aggregation cannot
recover information that the tests never reveal.

The final limitation is \emph{candidate generation}. On LiveCodeBench, even
after 500 attempts, $57\%$ of tasks still contain no correct candidate. For
those tasks, neither better tests nor a better selector can help. These three
failure modes suggest different remedies: improve the selection rule when the
available evidence is being used poorly, improve the generated tests when they
fail to separate candidates, and improve the generator when correct solutions
never appear.
% -----------------------------------------------------------------------------
\subsection{Effect of test-writer scale}
\label{app:test-writer-scale}

A natural question is whether the generated tests are weak simply because the
test-writing model is too small. We test this at $K=500$ while holding the
candidate programs fixed. Replacing the 7B test writer with a 32B model changes
CodeT accuracy by only $0.5$ percentage points on MBPP+ and $1.2$ points on
HumanEval+. The corresponding upper bound obtained from the observed test
outcomes changes by only $1.0$ and $0.3$ points. A larger test writer therefore
does little to close the verification gap in these experiments. The main
problem appears to be that the generated tests omit behaviors needed to
distinguish some incorrect programs from correct ones.

We omit LiveCodeBench from this comparison because the stored $K=500$ data do
not retain the candidate programs. Regenerating them would change the candidate
set and would no longer give a paired comparison.

% -----------------------------------------------------------------------------
\subsection{Usability of model-generated tests}
\label{app:selftest-coverage}

Before asking whether generated tests are informative, they must first be
generated in a form that can be parsed and executed. This is rarely a problem
for the native Qwen models on MBPP+ and HumanEval+, but it becomes important
for some compact models and on LiveCodeBench. Table~\ref{tab:selftest-coverage}
reports the fraction of tasks with usable generated tests and the average
number of tests produced.

\begin{table*}[!t]
\centering
\scriptsize
\setlength{\tabcolsep}{4.0pt}
\renewcommand{\arraystretch}{1.05}

\caption{\textbf{The ability to produce usable generated tests varies
substantially across model families.}
``Usable'' is the fraction of tasks in the generated-test pool for which the
tests can be parsed and executed. ``Tests'' is the mean number of generated
tests.}
\label{tab:selftest-coverage}

\resizebox{\textwidth}{!}{%
\begin{tabular}{@{}l rr rr rr@{}}
\toprule
& \multicolumn{2}{c}{\textbf{MBPP+}}
& \multicolumn{2}{c}{\textbf{HumanEval+}}
& \multicolumn{2}{c}{\textbf{LiveCodeBench}} \\
\cmidrule(lr){2-3}
\cmidrule(lr){4-5}
\cmidrule(l){6-7}

Checkpoint
& Usable & Tests
& Usable & Tests
& Usable & Tests \\
\midrule

\multicolumn{7}{@{}l}{\textit{Native Qwen2.5-Coder}} \\
Coder-1.5B & 0.99 & 10.8 & 0.99 & 11.2 & 0.95 & 8.4 \\
Coder-3B   & 1.00 & 11.5 & 1.00 & 11.7 & 0.98 & 9.8 \\
Coder-7B   & 1.00 & 11.8 & 1.00 & 11.8 & 0.95 & 9.5 \\
Coder-14B  & 1.00 & 12.0 & 1.00 & 11.9 & 0.89 & 8.9 \\
Coder-32B  & 1.00 & 12.0 & 1.00 & 12.0 & 0.84 & 8.4 \\

\addlinespace[3pt]
\multicolumn{7}{@{}l}{\textit{Quantized Qwen2.5-Coder}} \\
Coder-3B AWQ4  & 1.00 & 11.4 & 1.00 & 11.5 & 0.97 & 9.7 \\
Coder-3B GPTQ4 & 1.00 & 11.3 & 1.00 & 11.7 & 0.99 & 9.9 \\
Coder-3B GPTQ8 & 1.00 & 11.4 & 1.00 & 11.7 & 0.96 & 9.6 \\
Coder-7B AWQ4  & 1.00 & 11.6 & 1.00 & 11.8 & 0.98 & 9.8 \\
Coder-7B GPTQ4 & 1.00 & 11.6 & 1.00 & 11.7 & 0.97 & 9.6 \\
Coder-7B GPTQ8 & 1.00 & 11.7 & 1.00 & 11.8 & 0.97 & 9.7 \\

\addlinespace[3pt]
\multicolumn{7}{@{}l}{\textit{Other model families}} \\
Granite-Code-3B & 0.97 & 10.0 & 0.98 & 10.4 & 0.84 & 5.4 \\
Minitron-8B     & 1.00 & 11.3 & 1.00 & 11.5 & 0.61 & 6.0 \\
R1-Distill-1.5B & 0.75 & 8.0  & 0.69 & 7.3  & 0.00 & 0.0 \\

\bottomrule
\end{tabular}}
\end{table*}

Table~\ref{tab:selftest-coverage} shows that native and quantized Qwen models
almost always produce usable tests on MBPP+ and HumanEval+. The differences
become much larger across model families and on LiveCodeBench. For example,
R1-Distill-1.5B produces usable tests for only $75\%$ of MBPP+ tasks and
$69\%$ of HumanEval+ tasks, and none of its LiveCodeBench test generations are
usable. Minitron-8B is reliable on the EvalPlus benchmarks but falls to
$61\%$ usability on LiveCodeBench. These failures occur before we can even ask
whether the tests are good enough to distinguish correct programs from
incorrect ones.

The controlled Qwen3.6/Bonsai LiveCodeBench experiment uses the compact
line-based format described in Section~\ref{app:bonsai}. Usable tests are
produced for $97.7\%$, $51.4\%$, and $75.4\%$ of tasks by Q1\_0, AWQ4, and
bf16, respectively. We report these differences explicitly because generating
a usable test suite is itself part of the self-verification problem.
% =============================================================================
\section{Measurement construction and validation}
\label{app:repro}
% =============================================================================

The results in this paper combine generated candidates, benchmark execution,
model-generated tests, checkpoint files, and systems measurements. Because
these pieces are produced separately, the comparisons depend on matching the
right candidates, tasks, and measurements. This section describes the task
populations and memory accounting used throughout the paper.

\subsection{Task populations, candidate pools, and memory accounting}

The main accuracy experiments use 50 candidates per task, while the deeper verification study uses separate Qwen2.5-Coder-7B pools with up to 500 candidates per task. Candidate correctness is always evaluated with tests kept separate from the evidence available to the selector. EvalPlus uses the full configured HumanEval+ and MBPP+ task sets, while LiveCodeBench uses the 175 problems from the stated \texttt{release\_v5} date range.

BigCodeBench uses the filtered population defined in Section~\ref{sec:setup}. The separate Qwen3.6/Bonsai study begins from a 380-task subset and retains 339 tasks after the same reference-solution check. The $K=500$ verification study and the MoE experiments use their own fixed task populations.

Checkpoint memory is measured directly from the model files used in each experiment. For example, the native Qwen2.5-Coder-32B checkpoint occupies $65.53\GB$. We report the footprint of the checkpoint actually used rather than inferring a hypothetical lower-precision footprint. The additional Qwen2.5-72B result is reported only as an accuracy comparison because we do not have a corresponding measured checkpoint footprint.

{\bf Model-memory calculation.\enspace}
Let $W_q$ denote checkpoint weight memory, $b_q$ the number of KV-cache bytes
per token, $P_x$ the prompt length, and $C_{x,i}$ the completion length of
candidate $i$. For $K$ candidates from the same prompt, we count the shared
prompt once and each completion separately:
\begin{equation}
V_q(x,K)
=
b_q\!\left(P_x+\sum_{i=1}^{K}C_{x,i}\right)
=
b_q\!\left(P_x+K\bar C_x\right).
\label{eq:kv-memory}
\end{equation}
The resulting model memory is
\[
A_q(x,K)=W_q+V_q(x,K).
\]
This quantity includes the checkpoint weights and KV cache, but not activations,
driver state, allocator overhead, or serving-engine workspace. Measured process
memory is therefore reported separately.

Table~\ref{tab:kv-regimes} shows how the KV cache changes the memory footprint
as the retry budget increases.

\begin{table*}[t]
\centering
\scriptsize
\setlength{\tabcolsep}{2.2pt}
\renewcommand{\arraystretch}{1.04}

\caption{\textbf{Checkpoint weights, KV cache, and accuracy across retry budgets.}
$W$ is checkpoint weight memory, $V$ is the KV cache from
Equation~\ref{eq:kv-memory}, and $A=W+V$ is their sum. The table contains the
configurations used for the main accuracy--memory curves in
Figure~\ref{fig:accuracy-memory-concrete}. Memory values exclude runtime
overhead; measured process-memory results are reported separately in
Table~\ref{tab:memory-calibration}.}
\label{tab:kv-regimes}

\resizebox{\textwidth}{!}{%
\begin{tabular}{@{}ll r rrrr rrrr rrrr@{}}
\toprule
& & & \multicolumn{4}{c}{KV cache $V$ (GB)}
& \multicolumn{4}{c}{Weights + KV $A$ (GB)}
& \multicolumn{4}{c}{Accuracy (\%)} \\
\cmidrule(lr){4-7}
\cmidrule(lr){8-11}
\cmidrule(l){12-15}

Benchmark & Checkpoint & $W$ (GB)
& $K{=}1$ & $K{=}4$ & $K{=}8$ & $K{=}50$
& $K{=}1$ & $K{=}4$ & $K{=}8$ & $K{=}50$
& $K{=}1$ & $K{=}4$ & $K{=}8$ & $K{=}50$ \\
\midrule

\multirow{4}{*}{MBPP+}
& Coder-3B AWQ4  &  2.69 & 0.01 & 0.04 & 0.07 & 0.45 &  2.70 &  2.73 &  2.76 &  3.14 & 62.5\% & 72.5\% & 75.0\% & 78.0\% \\
& Coder-7B AWQ4  &  5.57 & 0.02 & 0.07 & 0.13 & 0.78 &  5.59 &  5.64 &  5.70 &  6.35 & 71.7\% & 77.6\% & 78.6\% & 79.9\% \\
& Coder-7B bf16  & 15.23 & 0.02 & 0.07 & 0.13 & 0.78 & 15.25 & 15.30 & 15.36 & 16.01 & 69.6\% & 77.0\% & 78.0\% & 78.6\% \\
& Coder-32B bf16 & 65.53 & 0.09 & 0.29 & 0.56 & 3.41 & 65.62 & 65.82 & 66.09 & 68.94 & 77.9\% & 80.8\% & 81.8\% & 82.9\% \\

\midrule
\multirow{4}{*}{HumanEval+}
& Coder-3B AWQ4  &  2.69 & 0.02 & 0.07 & 0.13 & 0.80 &  2.71 &  2.76 &  2.82 &  3.49 & 75.8\% & 86.1\% & 88.5\% & 90.9\% \\
& Coder-7B AWQ4  &  5.57 & 0.02 & 0.07 & 0.13 & 0.75 &  5.59 &  5.64 &  5.70 &  6.32 & 80.0\% & 85.7\% & 86.7\% & 87.8\% \\
& Coder-7B bf16  & 15.23 & 0.02 & 0.07 & 0.13 & 0.75 & 15.25 & 15.30 & 15.36 & 15.98 & 82.5\% & 87.8\% & 89.4\% & 90.9\% \\
& Coder-32B bf16 & 65.53 & 0.16 & 0.49 & 0.94 & 5.61 & 65.69 & 66.02 & 66.47 & 71.14 & 85.3\% & 88.1\% & 89.0\% & 89.3\% \\

\midrule
\multirow{4}{*}{LiveCodeBench}
& Coder-3B AWQ4  &  2.69 & 0.04 & 0.08 & 0.14 & 0.80 &  2.73 &  2.77 &  2.83 &  3.49 &  8.7\% & 15.2\% & 18.4\% & 26.3\% \\
& Coder-7B AWQ4  &  5.57 & 0.05 & 0.09 & 0.15 & 0.80 &  5.62 &  5.66 &  5.72 &  6.37 & 16.1\% & 23.6\% & 26.6\% & 31.4\% \\
& Coder-7B bf16  & 15.23 & 0.05 & 0.09 & 0.15 & 0.80 & 15.28 & 15.32 & 15.38 & 16.03 & 16.4\% & 24.2\% & 27.7\% & 33.7\% \\
& Coder-32B bf16 & 65.53 & 0.33 & 0.89 & 1.63 & 9.46 & 65.86 & 66.42 & 67.16 & 74.99 & 27.1\% & 33.2\% & 35.1\% & 38.3\% \\

\midrule
\multirow{4}{*}{BigCodeBench}
& Coder-3B AWQ4  &  2.69 & 0.03 & 0.07 & 0.13 & 0.73 &  2.72 &  2.76 &  2.82 &  3.42 & 39.4\% & 53.1\% & 57.5\% & 64.0\% \\
& Coder-7B AWQ4  &  5.57 & 0.03 & 0.08 & 0.15 & 0.82 &  5.60 &  5.65 &  5.72 &  6.39 & 46.4\% & 58.4\% & 62.0\% & 67.9\% \\
& Coder-7B bf16  & 15.23 & 0.03 & 0.08 & 0.15 & 0.82 & 15.26 & 15.31 & 15.38 & 16.05 & 48.0\% & 59.4\% & 62.7\% & 67.8\% \\
& Coder-32B bf16 & 65.53 & 0.14 & 0.30 & 0.52 & 2.79 & 65.67 & 65.83 & 66.05 & 68.32 & 56.6\% & 64.9\% & 67.1\% & 69.7\% \\

\bottomrule
\end{tabular}}
\end{table*}

Table~\ref{tab:kv-regimes} shows why moderate retry changes memory much less
than changing checkpoint size. On MBPP+, for example, Coder-7B AWQ4 at $K=8$
uses $5.57\GB$ for its weights and only $0.13\GB$ for the KV cache, for a total
of $5.70\GB$. Coder-32B bf16 at $K=1$ uses $65.53\GB$ of weights and
$0.09\GB$ of KV cache. The additional memory needed for several attempts is
therefore small compared with the difference in checkpoint size. %For BigCodeBench, the accuracy trajectory in Figure~\ref{fig:accuracy-memory-concrete} uses the canonical $1{,}034$-task population, while its memory coordinate uses the fixed systems workload used for the serving measurements.

At much higher concurrency, KV memory becomes more important. For example, the
60-prompt MBPP+ workload for Coder-7B AWQ4 at $K=50$ corresponds to
$46.9\GB$ of KV state if all requests are summed. This is a logical cache
requirement rather than a measured process peak: vLLM allocates and schedules
KV memory internally. We therefore keep the weights-plus-KV calculation
separate from the process-memory measurements reported next.
% -----------------------------------------------------------------------------
\subsection{From model memory to measured process memory}
\label{app:memory-calibration}

The weights-plus-KV calculation isolates the memory that changes directly with
checkpoint choice and retry depth, but a running server also uses memory for
runtime state, allocator overhead, and serving-engine workspace. We therefore
measure peak process memory on the same B200/vLLM stack.

For each configuration, we gradually reduce
\texttt{gpu\_memory\_utilization} until the run no longer succeeds and report
the lowest successful setting. This gives a measured process-level check on the
weights-plus-KV calculation; it is not intended to identify the best serving
configuration.

\begin{table}[t]
\centering
\scriptsize
\setlength{\tabcolsep}{4.2pt}
\renewcommand{\arraystretch}{1.05}

\caption{\textbf{Measured process memory for representative B200/vLLM configurations.}
For each configuration, we report the lowest tested
\texttt{gpu\_memory\_utilization} setting that completed successfully and the
corresponding peak process memory. The utilization setting controls vLLM's
cache allocation and is not itself a device-memory limit.}
\label{tab:memory-calibration}

\begin{tabular}{@{}lrrr@{}}
\toprule
Configuration
& \shortstack{Lowest successful\\utilization}
& \shortstack{Peak\\(GB)}
& \shortstack{Process-memory\\saving} \\
\midrule
Coder-7B AWQ4, $K=8$   & 0.045 & 11.55 & $6.32\times$ \\
Coder-7B AWQ4, $K=50$  & 0.046 & 11.73 & $6.23\times$ \\
Coder-7B bf16, $K=8$   & 0.098 & 21.58 & $3.38\times$ \\
Coder-32B bf16, $K=1$  & 0.370 & 73.02 & $1.00\times$ \\
\bottomrule
\end{tabular}
\end{table}

Table~\ref{tab:memory-calibration} shows the same ordering as the
weights-plus-KV calculation, although the savings are smaller once runtime
overhead is included. Coder-7B bf16 at $K=8$ peaks at $21.58\GB$, compared
with $73.02\GB$ for Coder-32B bf16 at $K=1$, giving a $3.38\times$ reduction
in measured process memory. The corresponding weights-plus-KV comparison is
about $4.3\times$. Coder-7B AWQ4 at $K=50$ peaks at $11.73\GB$, or
$6.23\times$ less than the 32B reference.

The small difference between the two AWQ4 measurements should not be
interpreted as a precise effect of retry depth. Their measured peaks differ by
only $0.18\GB$, while adjacent settings in the sweep correspond to roughly
$0.38\GB$ of process memory. We therefore use the process measurements to
validate the overall scale of the memory savings and use the weights-plus-KV
calculation to describe how memory changes with $K$.

The accuracy results were not regenerated at these reduced memory settings.
They come from the standard candidate pools, generated with
\texttt{gpu\_memory\_utilization}=0.9. The memory sweep is used only to measure
the physical process footprint of configurations that successfully execute,
not to produce a separate set of accuracy results.

% -----------------------------------------------------------------------------
\subsection{Runtime and energy measurement protocol}

The main Qwen2.5-Coder systems experiments use vLLM on B200 GPUs
\citep{kwon2023vllm}. At each retry budget, a fixed set of 40--60 prompts is
run concurrently, and wall time is measured until the entire workload
finishes. This reflects the deployment setting we care about: multiple
candidate streams can run in parallel and use hardware that would otherwise be
underutilized by single-stream decoding.

The Qwen3.6/Bonsai experiments and the dense Apple model-size comparisons use
llama.cpp for every checkpoint within a comparison. We never form a speedup by
comparing a llama.cpp run with a vLLM run. All Apple measurements use the same
llama.cpp source revision,
\texttt{4df29be4f4c3673f428170fda944a5b19f743bb8}
(release \texttt{b10454}).

Within each systems experiment, prompts and generation limits are held fixed.
Reported wall times are based on at least three repetitions. For the
HumanEval+ recovery comparison, we also run both checkpoint orders, with a
discarded warm-up followed by five measured repetitions per checkpoint.
Completion-token counts vary by only $0.9$--$1.8\%$ across these blocks,
whereas wall time varies by $16.0$--$28.9\%$. We therefore report latency
ratios at modest precision. Runtime is always measured directly for the
checkpoint being served; we do not infer the speed of a quantized model from
the corresponding native model.

% -----------------------------------------------------------------------------
\subsection{Scope and limitations}
\label{app:limitations}

All experiments are in code generation, where executable tests provide a
particularly clear notion of correctness. The results therefore establish a
memory--retry tradeoff for tasks with executable feedback; they do not imply
that the same tradeoff will hold for open-ended generation or other settings
where correctness cannot be checked directly.

The runtime and energy measurements use public-test selection. A fully
end-to-end system based on model-generated tests would also pay the cost of
generating and executing those tests, which is not included in the systems
numbers reported here. Similarly, the B200 process-memory measurements are
separate calibration runs rather than the runs used to generate the accuracy
pools.

The systems results are also specific to the hardware and serving stack. For
the one-bit Qwen3.6 comparison, the same $K=4$ deployment is slower than bf16
on M3 Max and faster on M5 Pro. In the dense-model comparison, every measured
recovery point at $K\leq4$ is favorable in wall time and energy on both chips.
At $K=8$ on MBPP+, M3 Max is unfavorable, while the two M5 Pro wall-time
measurements straddle break-even and both energy measurements remain
unfavorable. At $K=16$ on LiveCodeBench, both wall time and energy are
unfavorable on both chips, although the M5 Pro ratios are closer to break-even.
These observations should not be interpreted as a universal retry threshold;
the break-even depends on the model, workload, hardware, and serving stack.

Apple energy is reported only when every expected batch has a power record,
sampling covers at least half of each batch, and the measurement passes the
same consistency checks used throughout the study. Some reported device
results combine compatible disjoint shards; within each shard, the reference
and retried configurations are measured on the same physical device.

Adaptive stopping could reduce the average number of attempts, but it would
introduce sequential decision rounds and therefore represents a different
serving strategy from the fixed parallel retry budgets studied here.

% -----------------------------------------------------------------------------
\subsection{Validation of candidate and verifier data}

{\bf Matching candidates to generated tests.\enspace} A generated-test result is meaningful only if it is matched to the candidate
programs on which those tests were actually executed. Matching task names or
filenames is not sufficient: if two candidate generations differ, combining
correctness labels from one with test outcomes from the other can produce
invalid selector results. We therefore require the candidates, correctness
labels, mean log-probabilities, and candidate-by-test outcomes to come from the
same generation before including a CodeT result in a paired analysis.

Under this rule, the pooled CodeT improvement reported in
Section~\ref{sec:codet-delivery} contains 58 verified configurations:
24 on MBPP+, 21 on HumanEval+, and 13 on BigCodeBench. We exclude 17
LiveCodeBench configurations from that pooled statistic because their
generated-test experiments intentionally use independently generated candidate
pools. Those experiments remain valid for comparisons within their own pools,
but we do not pair them with a different generation.

Four additional identifiers correspond to duplicate generations with
materially different correctness labels: DeepSeek-6.7B, DeepSeek-33B, and
DeepSeek-1.3B on HumanEval+, and Magicoder on MBPP+. We exclude these from
paired analyses. DeepSeek-6.7B and Magicoder also show unusually high rates of
malformed outputs. These exclusions do not affect the controlled Qwen
precision comparisons, the native accuracy--time experiments, the fallback
analysis, the memory results, or the $K=500$ Qwen study.

{\bf Invalid candidate pools.\enspace}
We never exclude a candidate pool simply because its score is low. A pool is
removed only when an independent check identifies a systematic generation,
extraction, or execution failure.

{\bf Generated-test failures.\enspace}
When a task produces no usable generated tests, CodeT falls back to the
highest-confidence candidate rather than dropping the task. Test-generation
failures therefore remain part of the reported accuracy. The native
LiveCodeBench generated-test experiments use independently generated candidate
pools, so comparisons with the separate public-test runs are descriptive
rather than paired.

% {\bf Usable generated tests and matched baselines.\enspace}
% When model-generated tests are used, we distinguish between whether a task in
% the generated-test experiment produced usable test outcomes and whether the
% corresponding task is present in the main candidate population. For EvalPlus,
% the reported generated-test configurations cover the full candidate
% population. The native LiveCodeBench generated-test experiments use independent
% candidate pools, so there is no task-by-task pairing with the separate
% public-test generation.

% For BigCodeBench, generated-test coverage is $100.0\%$ for the four 3B/7B
% GPTQ matrices, $99.9\%$ for matrices that omit task 348, and $99.7\%$ for
% Coder-3B AWQ4. In the controlled Qwen3.6 experiment, usable generated tests are
% available for $97.7\%/51.4\%/75.4\%$ of LiveCodeBench tasks and
% $83.5\%/93.5\%/96.2\%$ of BigCodeBench tasks for
% Q1\_0/AWQ4/bf16, respectively.

% Whenever CodeT is evaluated on a subset of tasks, its one-shot baseline is
% computed on that same subset. We do not compare a subset-conditioned CodeT
% result inferentially with a full-population one-shot result.

% {\bf Common task sets.\enspace}
% Within a direct comparison, all quantities use the same valid task set whenever
% possible. When a figure directly compares public-test selection, candidate
% coverage, and CodeT, we use the same task set for all three.

{\bf Extraction and execution checks.\enspace}
CodeT requires the full candidate-by-test outcome matrix; a stored count of
passed tests is not enough. Before accepting a new model family, we check code
extraction, code fences, tokenizer-specific decoding, parse rates, expected
outputs, and preservation of line breaks. Program correctness is determined
from the expected output rather than process exit status alone.

For Qwen3.6/Bonsai on BigCodeBench, each candidate runs with a 16-GiB memory
limit. A pathological program can therefore fail without interrupting the rest
of the evaluation. We also verify candidate identities and the fixed 339-task
population before accepting the resulting verification measurements.

The native Qwen2.5-Coder LiveCodeBench generated-test experiments create their
own candidate solutions. Their CodeT results can therefore be compared across
checkpoints, but they are not paired candidate-by-candidate with the separate
public-test generation. A per-pool \texttt{PROVENANCE.json} file identifies
the intended candidate-by-test matrix explicitly. The reported 3B and 32B
LiveCodeBench CodeT accuracies are $18.3\%$ and $32.0\%$.

The controlled Qwen3.6 LiveCodeBench experiment instead executes generated
tests against the existing $K=50$ candidate pools. Each checkpoint contains
exactly 8,750 candidates over 175 tasks, with no missing candidate IDs,
log-probability mismatches, or correctness-label mismatches. Differences in
test usability therefore come from test generation and parsing rather than
from mismatched candidate pools.

Finally, within each $K=500$ experiment, every curve uses the same candidates,
generated tests, and candidate-by-test outcomes. CodeT and the upper bound
based on test-outcome patterns use independent Monte Carlo subsets, so their
plotted difference includes sampling variation from both estimates.
Independent 7B generations differ in one-shot accuracy by roughly
$1.5$--$2.3$ percentage points, which sets a practical scale below which
cross-generation differences should not be overinterpreted.

\end{document}